\documentclass[runningheads]{llncs}
\usepackage{graphicx}
\usepackage{amsmath,amssymb} % define this before the line numbering.
\usepackage{color}
\usepackage{subfig}
\usepackage{comment}
\usepackage{bm}
\usepackage{makecell}
\usepackage{wrapfig}
\usepackage{multirow}
\usepackage{makeidx}
\usepackage[hang,flushmargin]{footmisc}

\newcommand\blfootnote[1]{%
  \begingroup
  \renewcommand\thefootnote{}\footnote{#1}%
  \addtocounter{footnote}{-1}%
  \endgroup
}

\begin{document}
\title{Perturbation Robust Representations of Topological Persistence Diagrams} 
% Replace with your title

\titlerunning{Perturbation Robust Representations of Topological Persistence Diagrams}
% Replace with a meaningful short version of your title

\author{Anirudh Som\inst{1}\textsuperscript{*},
Kowshik Thopalli\inst{1}\textsuperscript{*},
Karthikeyan Natesan Ramamurthy\inst{2},\\ Vinay Venkataraman\inst{1}, Ankita Shukla\inst{3},\\ Pavan Turaga\inst{1}}

\index{Ramamurthy, Karthikeyan}

%\author{First Author\inst{1}\orcidID{0000-1111-2222-3333} \and
%Second Author\inst{2,3}\orcidID{1111-2222-3333-4444} \and
%Third Author\inst{3}\orcidID{2222--3333-4444-5555}}
%Please write out author names in full in the paper, i.e. full given and family names. 
%If any authors have names that can be parsed into FirstName LastName in multiple ways, please include the correct parsing, in a comment to the volume editors:
%\index{Lastnames, Firstnames}
%(Do not uncomment it, because you may introduce extra index items if you do that, we will use scripts for introducing index entries...)
\authorrunning{A. Som \textit{et al.}}
% Replace with shorter version of the author list. If there are more authors than fits a line, please use A. Author et al.
%

\institute{Geometric Media Lab, Arizona State University \\
\email{ \{asom2,kthopall,vvenka18,pturaga\}@asu.edu}
\and
IBM T. J. Watson Research Center, 
\email{knatesa@us.ibm.com}\\
%\url{http://www.springer.com/gp/computer-science/lncs} 
\and
Indraprastha Institute of Information Technology-Delhi,
\email{ankitas@iiitd.ac.in}
}

%\institute{Princeton University, Princeton NJ 08544, USA \and
%Springer Heidelberg, Tiergartenstr. 17, 69121 Heidelberg, Germany
\maketitle              % typeset the header of the contribution
\begin{abstract}
Topological methods for data analysis present opportunities for enforcing certain invariances of broad interest in computer vision, including view-point in activity analysis, articulation in shape analysis, and measurement invariance in non-linear dynamical modeling. The increasing success of these methods is attributed to the complementary information that topology provides, as well as  availability of tools for computing topological summaries such as persistence diagrams. However, persistence diagrams are multi-sets of points and hence it is not straightforward to fuse them with features used for contemporary machine learning tools like deep-nets. In this paper we present theoretically well-grounded approaches to develop novel perturbation robust topological representations, with the long-term view of making them amenable to fusion with contemporary learning architectures. We term the proposed representation as Perturbed Topological Signatures, which live on a Grassmann manifold and hence can be efficiently used in machine learning pipelines. We explore the use of the proposed descriptor on three applications: 3D shape analysis, view-invariant activity analysis, and non-linear dynamical modeling. We show favorable results in both high-level recognition performance and time-complexity when compared to other baseline methods.
\keywords{Invariance Learning, Topological Data Analysis, Persistence Diagrams, Grassmann Manifold, Perturbed Topological Signature.}
\end{abstract}

\blfootnote{\noindent This work was supported in part by ARO grant number W911NF-17-1-0293 and NSF CAREER award 1452163.  \textsuperscript{*}The first two authors contributed equally.}

\section{Introduction}\label{intro_section}

Over the years, tools from topological data analysis (TDA) have been used to characterize the invariant structure of data obtained from a noisy sampling of an underlying metric space \cite{edelsbrunner2010computational}. Invariance learning is a fundamental problem in computer vision, since common transformations can diminish the performance of algorithms significantly. Past work in invariance learning has fallen into one of two classes. The first approach involves ad-hoc choices of features or metrics between features that offer some invariance to specific factors \cite{begelfor2006affine}. However, this approach has suffered due to lack of generalizable solutions. The other approach is to increase the training size by collecting samples that capture all the variations of the data, so that the learning algorithm can implicitly marginalize out the variations. A similar effect can be achieved via simple data augmentation \cite{rahmani2017learning}. 

\setlength{\intextsep}{0pt}%
\begin{figure*}[t!]
\centering
\includegraphics[width=1\linewidth]{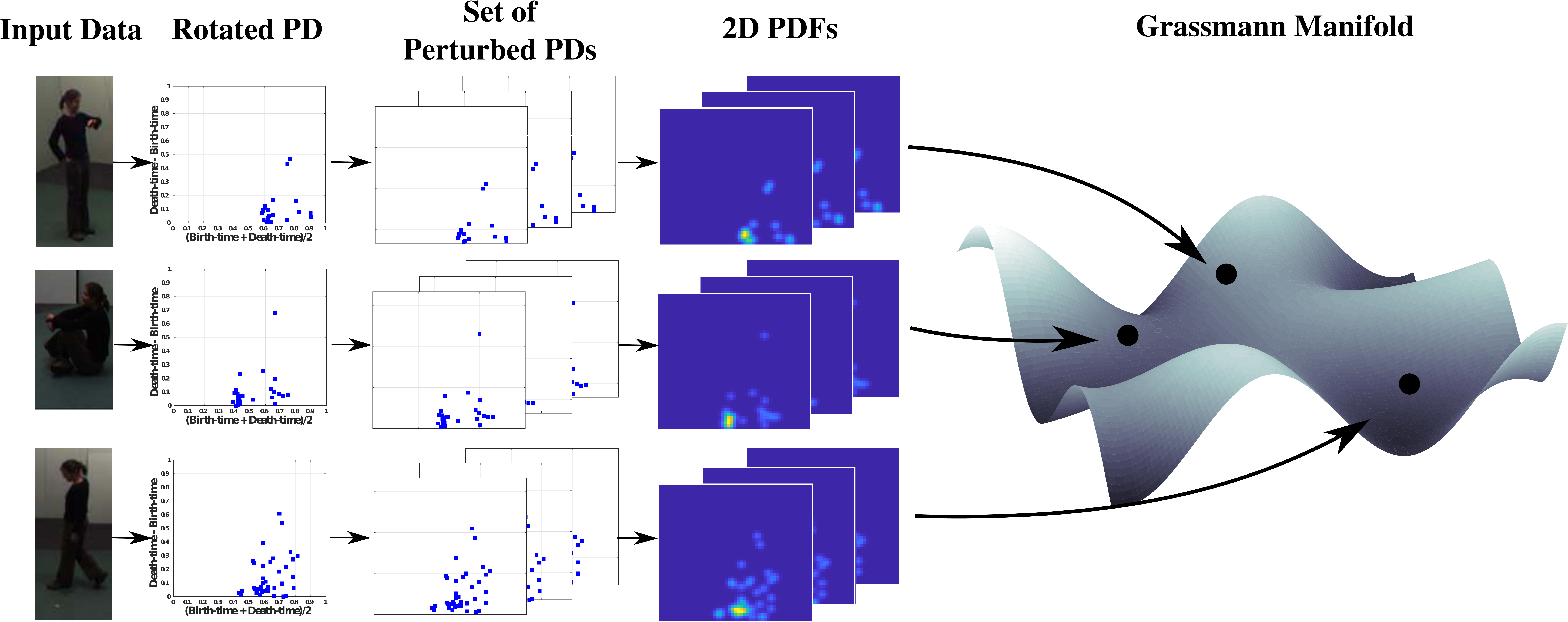}
	\caption{Illustration of the sequence of steps leading to the proposed Perturbed Topological Signature (PTS) representation. For a given input dataset, the PDs are computed and transformed to maximally occupy the 2D space. A set of perturbed PDs is created, with each perturbed PD having its points displaced by a certain amount about its initial position. For each PD in the set, a 2D PDF is constructed using a Gaussian kernel function via kernel density estimation. The set of 2D PDFs capture a wide range of topological noise for the given input data and are summarized using a subspace structure, equivalent to a point on the Grassmann manifold.}\label{framework_pipeline}
\end{figure*}

In this context, TDA has emerged as a surprisingly powerful tool to analyze underlying invariant properties of data before any contextual modeling assumptions or the need to extract actionable information kicks in. Generally speaking, TDA seeks to characterize the \emph{shape} of high dimensional data by quantifying various topological invariants such as connected components, cycles, high-dimensional holes, level-sets and monotonic regions of functions defined on the data \cite{edelsbrunner2010computational}. Topological invariants are those properties that do not change under smooth deformations like stretching, bending, and rotation, but without tearing or gluing surfaces. We illustrate the connections between topological invariants and learning invariant representations for vision via three applications:

\smallskip
\noindent {\bf 1) Point cloud shape analysis: } Shape analysis of 3-dimensional (3D) point cloud data is a topic of major current interest due to emergence of Light Detection and Ranging (LIDAR) based vision systems in autonomous vehicles. It has been a difficult problem to solve with contemporary methods (\textit{e.g.} deep learning) due to the non-vectorial nature of the representations. While there is interest in trying to extend deep-net architectures to point-cloud data \cite{scarselli2009graph,masci2015geodesic,yi2017syncspeccnn,monti2017geometric,hofer2017deep}, the invariance one seeks is that of {\em shape articulation}, \textit{i.e.} stretching, skewing, rotation of shape that does not alter the fundamental object class. These invariances are optimally defined in terms of topological invariants. 

\noindent {\bf 2) Video analysis:} One of the long-standing problems in video analysis, specific to human action recognition, is to deal with variation in {\em body type, execution style}, and {\em view-point} changes. Work in this area has shown that temporal self-similarity matrices (SSMs) are a robust feature and provide general invariance to the above factors \cite{junejo2011view}. Temporal self-similarities can be quantified by scalar field topological constructions defined over video features, leading to representations with encoded invariances not relying on brute-force training data.

\noindent {\bf 3) Non-linear dynamical modeling:} Many time-series analysis problems have been studied under the lens of non-linear dynamical modeling: including motion-capture analysis, wearable-based activity analysis \textit{etc}. Results from dynamical systems theory (Takens' embedding theorem \cite{takens1981detecting}) suggest that the placement-invariant property may be related to the topological properties of reconstructed dynamical attractors via delay-embeddings. 

One of the prominent TDA tools is persistent homology. It provides a multi-scale summary of different homological features \cite{Edelsbrunner2002}. This multi-scale information is represented using a persistence diagram (PD), a 2-dimensional (2D) Cartesian plane with a multi-set of points. For a point $(b,d)$ in the PD, a homological feature appears at scale $b$ and disappears at scale $d$. Due to the simplicity of PDs, there has been a surge of interest to use persistent homology for summarizing high-dimensional complex data and has resulted in its successful implementation in several research areas \cite{perea2015sliding,tralie2018quasi,chintakunta2015entropy,dabaghian2012topological,chung2009persistence,heath2010image,singh2008topological,venkataraman2016persistent}. However, application of machine learning (ML) techniques on the space of PDs has always been a challenging task. The gold-standard approach for measuring the distance between PDs is the \textit{Bottleneck} or the $p$\textit{-Wasserstein} metric \cite{mileyko2011probability,turner2014frechet}. However, a simple metric structure is not enough to use vector based ML tools such as support vector machines (SVMs), neural networks, random forests, decision trees, principal component analysis and so on. These metrics are only stable under small perturbations of the data which the PDs summarize, and the complexity of computing distances between PDs grows in the order of $\mathcal{O}(n^3)$, where $n$ is the number of points in the PD \cite{bertsekas1981new}. Efforts have been made to overcome these problems by attempting to map PDs to spaces that are more suitable for ML tools \cite{anirudh2016riemannian,bubenik2015statistical,rouse2015feature,pachauri2011topology,reininghaus2015stable,adams2017persistence}. A comparison of some recent algorithms for machine learning over topological descriptors can be found in \cite{Seversky2016}. More recently, topological methods have also shown early promise in improving performance of image-based classification algorithms in conjunction with deep-learning \cite{Dey2017}.

\noindent\textbf{Contributions:} Using a novel perturbation framework, we propose a topological representation of PDs called \textit{Perturbed Topological Signature} (PTS). To do this we first generate a set of perturbed PDs by randomly shifting the points in the original PD by a certain amount. A perturbed PD is analogous to extracting the PD from data that is subjected to topological noise. Next, we utilize a 2D probability density function (PDF) estimated by kernels on each of the perturbed PDs to generate a smooth functional representation. Finally, we simplify and summarize the end representation-space for the set of 2D PDFs to a point on the Grassmann manifold (a non-constantly curved manifold). The framework described above is illustrated in figure \ref{framework_pipeline}. We develop very efficient ML pipelines over these topological descriptors by leveraging the known metrics and statistical results on the Grassmann manifold. We also develop a stability proof of the Grassmannian representations w.r.t. the normalized geodesic distance over the Grassmannian and the {\em Wasserstein} metrics over PDs. Experiments show that our proposed framework recovers the lost performance due to functional methods, while still enjoying orders of magnitude faster processing times over the classical $p${\em-Wasserstein} and {\em Bottleneck} approaches.

\noindent\textbf{Outline of the paper:} Section \ref{preliminary_section} provides the necessary background on topological data analysis and the Grassmannian. Section \ref{related_work_section} discusses related work, while section \ref{proposed_method_section} describes the proposed framework and end representation of the PD for statistical learning tasks. Section \ref{experiments_section} describes the experiments and results. Section \ref{conclusion_section} concludes the paper.

%%%%%%%%%%%%%%%%%%%%%%%%%%%%%%%%%%%%%%%%
\section{Preliminaries} \label{preliminary_section}
%%%%%%%%%%%%%%%%%%%%%%%%%%%%%%%%%%%%%%%%

\noindent \textbf{Persistent Topology:} Consider a graph $\mathcal{G} = \{\mathcal{V}, \mathcal{E}\}$ on the high-dimensional point cloud, where $\mathcal{V}$ is the set of $|\mathcal{V}|$ nodes and $\mathcal{E}$ defines the neighborhood relations between the samples. To estimate the topological properties of the graph's shape, a simplicial complex $\mathcal{S}$ is constructed over $\mathcal{G}$. We denote $\mathcal{S}=(\mathcal{G},\Sigma)$, where $\Sigma$ is a family of non-empty level sets of $\mathcal{G}$, with each element $\sigma \in \Sigma$ is a simplex \cite{Edelsbrunner2002}. These simplices are constructed using the the $\epsilon$-neighborhood rule, $\epsilon$ being the scale parameter \cite{Edelsbrunner2002}.
In TDA, Betti numbers $\beta_i$ provide the rank of the homology group $H_i$. For instance, $\beta_0$ denotes the number of connected components, $\beta_1$ denotes the number of holes or loops, $\beta_2$ denotes the number of voids or trapped volumes, \textit{etc}. They provide a good summary of a shape's topological features. However, two shapes with same Betti numbers can have very different PDs since PDs summarize the birth vs death time information of each topological feature in a homology group. Birth time $(b)$ signifies the scale at which the group is formed and death time $(d)$ is the scale at which it ceases to exist. The difference between the death and the birth times is the lifetime of the homology group $l = |d-b|$. Each PD is a multiset of points $(b,d)$ in $\mathbb{R}^2$ and is hence represented graphically as a set of points in a 2D plane. The diagonal where $b=d$ is assumed to contain an infinite number of points since they correspond to groups of zero persistence.

We use the Vietoris-Rips (VR) construction denoted by VR($\mathcal{G}$, $\epsilon$) to obtain simplicial complexes from $\mathcal{G}$ for a given scale $\epsilon$ \cite{edelsbrunner2010computational}. An algorithm for computing homological persistence is provided in \cite{Edelsbrunner2002} and an efficient dual variant that uses co-homology is described in \cite{de2011dualities}. The VR construction obtains the topology of the distance function on the point cloud data. However, given a graph $\mathcal{G}$, and a function $g$ defined on the vertices, it is also possible to quantify the topology induced by $g$ on $\mathcal{G}$. For example, we may want to study the topology of the sub-level or super-level sets of $g$. This is referred to as \emph{scalar field topology} since $g: \mathcal{V} \rightarrow \mathbb{R}$. A well-known application of this in vision is in 3D shape data, where the graph $\mathcal{G}$ corresponds to the shape mesh and $g$ is a function, such as heat kernel signature (HKS) \cite{sun2009concise}, defined on the mesh \cite{li2014persistence}. The PD of the $H_0$ homology group of the super-level sets now describes the evolving segments of regions in the shape. For instance, if we compute the PD of the super-level sets induced by HKS in an \emph{octopus} shape, we can expect to see eight highly persistent segments corresponding to the eight legs. This is because the HKS values are high at regions of high curvature in the shape. In scalar field constructions, the PDs can be obtained efficiently using the Union-Find algorithm by first sorting the nodes of $\mathcal{G}$ as per their function magnitude and keeping a trail of the corresponding connected components \cite{cormen2001introduction}.

\smallskip
\noindent
\textbf{Distance Metrics between PDs:} PDs are invariant to rotations, translations and scaling of a given shape, and under continuous deformation conditions are invariant to slight permutations of the vertices \cite{cohen2007stability,cohen2010lipschitz}. The two classical metrics to measure distances between PDs \textit{X} and \textit{Y} are the \textit{Bottleneck} distance and the $p$-\textit{Wasserstein} metric \cite{mileyko2011probability,turner2014frechet}. They are appealing as they reflect any small changes such as perturbations of a measured phenomenon on the shape, which results in small shifts to the points in the persistence diagram. The {\em Bottleneck} distance is defined as
%\begin{equation}\label{bottleneck_eqn}
\small$d_{\infty}(X,Y) = \inf_{\eta: X \rightarrow Y} \sup_{x \in X} \| x-\eta(x) {\|}_\infty$,
%\end{equation}
\noindent with $\eta$ ranging over all bijections and $\|.\|_\infty$ is the $\infty$-norm. Equivalently, the $p$-{\em Wasserstein} distance is defined as 
%\begin{equation}\label{Wasserstein_q_eqn}
\small$d_{p}(X,Y) = ( \inf_{\eta: X \rightarrow Y} \sum_{x \in X} \| x-\eta(x) {\|}_\infty^p )^{1/p}$. 
%\end{equation}
\noindent However, the complexity of computing distances between PDs with $n$ points is $\mathcal{O}(n^3)$. These metrics also do not allow for easy computation of statistics and are unstable under large deformations \cite{bertsekas1981new}.

\smallskip
\noindent
\textbf{Grassmann Manifold:} Let $n, p$ be two positive integers such that $n>p>0$. The set of $p$-dimensional linear subspaces in $\mathbb{R}^n$ is called a Grassmann manifold, denoted by $\mathbb{G}_{p,n}$. Each point $\mathcal{Y}$ on $\mathbb{G}_{p,n}$ is represented as a basis, \textit{i.e.} a linear combination of the set of $p$ orthonormal vectors $Y_1, Y_2,\dots, Y_p$. The geometric properties of the Grassmannian have been used for various computer vision applications, such as object recognition, shape analysis, human activity modeling and classification, face and video-based recognition, \textit{etc} \cite{begelfor2006affine,hamm2008grassmann,turaga2011statistical,gopalan2012blur}. We refer our readers to the following papers that provide a good introduction to the geometry, statistical analysis, and techniques for solving optimization problems on the Grassmann manifold \cite{absil2004riemannian,edelman1998geometry,wong1967differential,chikuse2012statistics,absil2009optimization}. 

\smallskip
\noindent
\textbf{Distance Metrics between Grassmann Representations:} The minimal geodesic distance $(d_\mathbb{G})$ between two points $\mathcal{Y}_1$ and $\mathcal{Y}_2$ on the Grassmann manifold is the length of the shortest constant speed curve that connects these points. To do this, the velocity matrix $A_{\mathcal{Y}_1,\mathcal{Y}_2}$ or the inverse exponential map needs to be calculated, with the geodesic path starting at $\mathcal{Y}_1$ and ending at $\mathcal{Y}_2$. $A_{\mathcal{Y}_1,\mathcal{Y}_2}$ can be computed using the numerical approximation method described in \cite{liu2003optimal}. The geodesic distance between $\mathcal{Y}_1$ and $\mathcal{Y}_2$ is represented by the following equation: \small$d_\mathbb{G}(\mathcal{Y}_1,\mathcal{Y}_2) = trace(A_{\mathcal{Y}_1,\mathcal{Y}_2}{A_{\mathcal{Y}_1,\mathcal{Y}_2}}^\textrm{T})$ or \small$d_\mathbb{G}(\mathcal{Y}_1,\mathcal{Y}_2) = \sqrt[]{trace(\theta^T \theta)}$. Here $\theta$ is the principal angle matrix between $\mathcal{Y}_1, \mathcal{Y}_2$ and can be computed as $\theta = \textrm{arccos}(S)$, where $USV^T = \textrm{svd}(\mathcal{Y}_1^T \mathcal{Y}_2)$. To show the stability of the proposed PTS representations in section \ref{proposed_method_section}, we use the normalized geodesic distance represented by $d_\mathbb{NG}(\mathcal{Y}_1,\mathcal{Y}_2) = \frac{1}{D}d_\mathbb{G}(\mathcal{Y}_1,\mathcal{Y}_2)$, where $D$ is the maximum possible geodesic distance on $\mathbb{G}_{p,n}$ \cite{ji1987perturbation,li2014grassmann}.
The symmetric directional distance $(d_{\Delta})$ is another popular metric to compute distances between Grassmann representations with different $p$  \cite{sun2007further,wang2006subspace}. It is a widely used measure in areas like computer vision \cite{da2009normalized,basri2011approximate,bagherinia2011theory,luo2014video,yan2006general}, communications \cite{sharafuddin2010know}, and applied mathematics \cite{draper2014flag}. It is equivalent to the chordal metric \cite{ye2016schubert} and is defined as, $d_{\Delta}(\mathcal{Y}_1,\mathcal{Y}_2) = \big(\textrm{max}(k,l)-\sum_{i,j=1}^{k,l}({y_{1,i}}^{\textrm{T}}y_{2,j})^2\big)^{\frac{1}{2}}$. Here, $k$ and $l$ are subspace dimensions for the orthonormal matrices $\mathcal{Y}_1$ and $\mathcal{Y}_2$ respectively. For all our experiments, we restrict ourselves to distance computations between same-dimension subspaces, \textit{i.e.} $k=l$. The following papers propose methods to compute distances between subspaces of different dimensions \cite{sun2007further,wang2006subspace,ye2016schubert}.

%%%%%%%%%%%%%%%%%%%%%%%%%%%%%%%%%%%%%%%%
\section{Prior Art} \label{related_work_section}
%%%%%%%%%%%%%%%%%%%%%%%%%%%%%%%%%%%%%%%%

PDs provide a compact multi-scale summary of the different topological features. The traditional metrics used to measure the distance between PDs are the \textit{Bottleneck} and $p$-\textit{Wasserstein} metrics \cite{mileyko2011probability,turner2014frechet}. These measures are stable with respect to small continuous deformations of the topology of the inputs  \cite{cohen2007stability,cohen2010lipschitz}. However, they do poorly under large deformations. Further, a feature vector representation will be useful that is compatible with different ML tools that demand more than just a metric. To address this need, researchers have resorted to transforming PDs to other suitable representations \cite{anirudh2016riemannian,bubenik2015statistical,rouse2015feature,pachauri2011topology,reininghaus2015stable,adams2017persistence}. Bubenik proposed persistence landscapes (PL) which are stable and invertible functional representations of PDs in a Banach space \cite{bubenik2015statistical}. A PL is a sequence of envelope functions defined on the points in PDs that are ordered on the basis of their importance. Bubenik's main motivation for defining PLs was to derive a unique mean representation for a set of PDs which is not necessarily obtained using Fr\'echet means \cite{mileyko2011probability}. Their usefulness is however limited, as PLs can provide low importance to moderate size homological features that generally possess high discriminating power. 

Rouse \textit{et al.} create a simple vector representation by overlaying a grid on top of the PD and count the number of points that fall into each bin \cite{rouse2015feature}. This method is unstable, since a small shift in the points can result in a different feature representation. This idea has also appeared in other forms, some of which are described below. Pachauri \textit{et al.} transform PDs into smooth surfaces by fitting Gaussians centered at each point in the PD \cite{pachauri2011topology}. Reininghaus \textit{et al.} create stable representations by taking a weighted sum of positive Gaussians at each point above the diagonal and mirror the same below the diagonal but with negative Gaussians \cite{reininghaus2015stable}. Adams \textit{et al.} design   
persistence images (PI) by defining a regular grid and obtaining the integral of the Gaussian-surface representation over the bins defined on each grid \cite{adams2017persistence}. Both PIs and the multi-scale kernel defined by Reininghaus \textit{et al.} show stability with respect to the Wasserstein metrics and do well under small perturbations of the input data. They also weight the points using a weighting function, and this can be chosen based on the problem. Prioritizing points with medium lifetimes was used by Bendich \textit{et al.} to best identify the age of a human brain by studying its arterial geometry \cite{bendich2016persistent}. Cohen-Steiner \textit{et al.} suggested prioritizing points near the death-axis and away from the diagonal \cite{cohen2007stability}. 

In this paper, we propose a unique perturbation framework that overcomes the need for selecting a weighting function. We consider a range of topological noise realizations one could expect to see, by perturbing the points in the PD. We summarize the perturbed PDs by creating smooth surfaces from them and consider a subspace of these surfaces, which naturally becomes a point on the Grassmann manifold. We show the effectiveness of our features in section \ref{experiments_section} for different problems using data collected from different sensing devices. Compared to the $p$-\textit{Wasserstein} and \textit{Bottleneck} distances, the metrics defined on the Grassmannian are computationally less complex and the representations are independent of the number of points present in the PD. The proposed PTS representation is motivated from \cite{gopalan2012blur}, where the authors create a subspace representation of blurred faces and perform face recognition on the Grassmannian. Our framework also bears some similarities to \cite{anirudh2016riemannian}, where the authors use the square root representation of PDFs obtained from PDs.

%%%%%%%%%%%%%%%%%%%%%%%%%%%%%%%%%%%%%%%%
\section{Perturbed Topological Signatures} \label{proposed_method_section}
%%%%%%%%%%%%%%%%%%%%%%%%%%%%%%%%%%%%%%%%

In this section we go through details of each step in our framework's pipeline, illustrated in figure \ref{framework_pipeline}. In our experiments we transform the axes of the PD from $(b,d) \rightarrow (\frac{b+d}{2},d-b)$, with $b\leq d$.

\noindent \textbf{Create a set of Perturbed PDs:} We randomly perturb a given PD to create $m$ PDs. Each of the perturbed PDs has its points randomly displaced by a certain amount compared to the original. The set of randomly perturbed PDs retain the same topological information of the input data as the original PD, but together capture all the probable variations of the input data when subjected to topological noise. We constrain the extent of perturbation of the individual points in the PD to ensure that the topological structure of the data being analyzed is not abruptly changed. 

\noindent \textbf{Convert Perturbed PDs to 2D PDFs:} We transform the initial PD and its set of perturbed PDs to a set of 2D PDFs. We do this via kernel density estimation: by fitting a Gaussian kernel function with zero mean, standard deviation $\sigma$ at each point in the PD, and then normalizing the 2D surface. The obtained PDF surface is discretized over a $k\times k$ grid similar to the approach of Rouse \textit{et al.} \cite{rouse2015feature}. The standard deviation $\sigma$ (also known as bandwidth parameter) of the Gaussian is not known a priori and is fine-tuned to get best results. A multi-scale approach can also be employed by generating multiple surfaces using a range of different bandwidth parameters for each of the PDs and still obtain favorable results. Unlike other topological descriptors that use a weighting function over their functional representations of PDs \cite{reininghaus2015stable,adams2017persistence}, we give equal importance to each point in the PD and do not resort to any weighting function. Adams \textit{et al.} prove the stability of persistence surfaces obtained using general and Gaussian distributions ($\phi$), together with a weighting function ($f$), with respect to the $1$-{\em Wasserstein} distance between PDs in \cite[Thm. 4, 9]{adams2017persistence}. For Gaussian distributions, both $L_{1}$ and $L_{\infty}$ distances between persistence surfaces $\rho_B, \rho_{B'}$ are stable with respect to $1$-{\em Wasserstein} distance between PDs $B, B'$,
%\begin{equation}\label{stable_gaussian_persistence_surface}
$\| \rho_{B}-\rho_{B'} {\|}_1 \leq \ \sqrt[]{\frac{10}{\pi}} \ \frac{1}{\sigma} \ d_{1}(B,B')$.
%\end{equation}

\noindent \textbf{Projecting 2D PDFs to the Grassmannian:} 
Let $\rho(x,y)$ be an unperturbed persistence surface, and let $\rho(x + u, y + v)$ be a randomly shifted perturbation. Under assumptions of small perturbations, we have using Taylor's theorem:
\begin{align}
\begin{split}
\rho(x + u, y + v) - \rho(x, y) &\approx [\rho_x, \rho_y] [u , v]^T
\end{split}
\end{align}
\noindent Now, in the following, we interpret $\approx$ as an equality, enabling us to stack together the same equation for all $(x,y)$, to get a matrix-vector form
%\begin{align}\label{eq:perturbed1}
$\overline{\rho}_{pert}^{u,v} - \overline{\rho} = [\overline{\rho}_x, \overline{\rho}_y]_{N \times 2} [u , v]^T_{2 \times 1}$,
%\end{align}
\noindent where the overline indicates a discrete vectorization of the 2D functions. Here, $N$ is the total number of discretized samples from the $(x,y)$ plane. Now consider the set of all small perturbations of $\rho$, i.e. $span(\overline{\rho}_{pert}^{u,v} - \overline{\rho})$, over all $[u,v] \in \mathbb{R}^2$. It is easy to see that this set is just a 2D linear-subspace in $\mathbb{R}^N$ which coincides with the column-span of $[\overline{\rho}_x, \overline{\rho}_y]$. For a more general  affine-perturbation model, we can show that the required subspace corresponds to a 6-dimensional (6D) linear subspace, corresponding to the column-span of the $N \times 6$ matrix $[\rho_x, \rho_y, x\rho_x, x\rho_y, y\rho_x, y\rho_y]$. More details on this can be found in the supplement. In implementation, we perturb a given PD several times using random offsets, compute their persistence surfaces, use singular value decomposition (SVD) on the stacked matrix of perturbations, then select the $p$ largest left singular vectors, resulting in a $N \times p$ orthonormal matrix. Also, we vary the dimension of the subspace across a range of values. Since the linear span of our matrix can be further identified as a point on the Grassmann manifold, we adopt metrics defined over the Grassmannian to compare our perturbed topological signatures.

\noindent \textbf{Stability of Grassmannian metrics w.r.t. Wasserstein:} The next natural question to consider is whether the metrics over the Grassmannian for the perturbed stack are in any way related to the Wasserstein metric over the original PDs. Let the column span of $X = [\overline{\rho}_x, \overline{\rho}_y]$ be represented by $\mathcal{X}(\rho)$. Let $\rho_1, \rho_2$ be two persistence surfaces, then $\mathcal{X}(\rho_1), \mathcal{X}(\rho_2)$ are the subspaces spanned by $X_1 = [\overline{\rho}_{1,x}, \overline{\rho}_{1,y}]$ and $X_2 = [\overline{\rho}_{2,x}, \overline{\rho}_{2,y}]$ respectively. Following a result due to  Ji-Guang \cite{ji1987perturbation}, the normalized geodesic distance  $d_\mathbb{NG}$ between $\mathcal{X}_1$ and $\mathcal{X}_2$ is upper bounded as follows: $
%\begin{align}\label{Ji-guang_equation}
d_\mathbb{NG}(\mathcal{X}_1,\mathcal{X}_2) \leq \|X_1\|_F.\|X_1^\dagger\|_2.\frac{\|\Delta X\|_F}{\|X_1\|_F} = \|X_1^\dagger\|_2.\|\Delta X\|_F
%\end{align}
$. Here, $\|X^\dagger\|_2$ is the spectral norm of the pseudo-inverse of $X$, $\|X\|_F$ is the Frobenius norm, and $\Delta X = X_1 - X_2$. In the supplement, a full derivation is given, showing $
%\begin{align}
\|\Delta X\|_F^2 \leq {\frac{10}{\pi}} \ \frac{{2}}{\sigma^6} \ d_{1}^2(B_1,B_2) + 2\frac{\mathcal{K}^2}{\sigma^4}k_{max}^2N
%\end{align}
$, \noindent where $d_1(B_1,B_2)$ is the $1$-\textit{Wasserstein} metric between the original unperturbed PDs, $k_{max}$ is the maximum number of points in a given PD (a dataset dependent quantity), $N$ refers to the total number of discrete samples from $[0,1]^2$ and $\mathcal{K} = \frac{1}{(\sqrt[]{2\pi}\sigma)^2}$. This is the critical part of the stability proof. The remaining part requires us to upper bound the spectral norm $\|X^\dagger\|_2$. The spectral-norm of the pseudo-inverse of $X$, i.e. $\|X^\dagger\|_2$, is simply the inverse of the smallest singular-value of $X$, which in turn corresponds to the square-root of the smallest eigenvalue of $X^TX$. i.e. $\|X^\dagger\|_2 = \sigma_{max}(X^\dagger) = \frac{1}{\sigma_{min}(X)} = \frac{1}{\sqrt{\lambda_{min}(X^TX)}}.$

Given that $X = [\overline{\rho}_x, \overline{\rho}_y]$,  $X^TX$ becomes the 2D structure-tensor of a Gaussian mixture model (GMM). While we are not aware of any results that lower-bound the eigenvalues of a 2D GMMs structure-tensor, in the supplement we show an approach for 1D GMMs that indicates that the smallest eigenvalue can indeed be lower-bounded, if the standard-deviation $\sigma$ is upper-bounded. For example, a non-trivial lower-bound is derived for $\sigma < 1$ in the supplement. It is inversely proportional to the number of components in the GMM. We used $\sigma = 0.0004$ for all our experiments. The approach in the supplement is shown for 1D GMMs, and we posit that a similar approach applies for the 2D case, but it is cumbersome. In empirical tests, we find that even for 2D GMMs defined over the grid $[0,1]^2$, with $0< \sigma < 1$, the spectral-norms are always upper-bounded. In general, we find $\|X^\dagger\|_2 \leq k/\sqrt{g(\sigma)}$, where $g(\sigma)$ is a positive monotonically decreasing function of $\sigma$ in the domain $[0,1]$, and $k$ is the number of components in the GMM (points in a given PD). If we denote $k_{max}$ and $\sigma_{max}$ as the maximum allowable number of components in the GMM (max points in any PD in given database) and the maximum standard deviation respectively, an upper bound readily develops. Thus, we have
\begin{equation}
\text{\small $d_\mathbb{NG}(\mathcal{X}_1,\mathcal{X}_2) \leq \frac{k_{max}}{\sqrt{g(\sigma_{max}})}\sqrt{{\frac{10}{\pi}} \ \frac{{2}}{\sigma^6} \ d_{1}^2(B_1,B_2) + 2\frac{\mathcal{K}^2}{\sigma^4}k_{max}^2N}$}
\end{equation}

Please refer to the supplement for detailed derivation and explanation of the various constants in the above bound. We note that even though the above shows that the normalized Grassmannian geodesic distance over the perturbed topological signatures is stable w.r.t the $1$-\textit{Wasserstein} metric over PDs, it still relies on  knowledge of the maximum number of points in any given PD across the entire dataset $k_{max}$, and also on the sampling of the 2D grid. 

%%%%%%%%%%%%%%%%%%%%%%%%%%%%%%%%%%%%%%%%
%\clearpage
\section{Experiments} \label{experiments_section}
%%%%%%%%%%%%%%%%%%%%%%%%%%%%%%%%%%%%%%%%

In this section we first show the robustness of the PTS descriptor to different levels of topological noise using a sample of shapes from the SHREC 2010 dataset \cite{lian2010shrec}. We then test the proposed framework on three publicly available datasets: SHREC 2010 shape retrieval dataset \cite{lian2010shrec}, IXMAS multi-view video action dataset \cite{weinland2007action} and motion capture dataset \cite{ali2007chaotic}. We briefly go over the details of each dataset, and describe the experimental objectives and procedures followed. Finally, we show the computational benefits of comparing different PTS representations using the $d_{\mathbb{G}}$ and $d_{\Delta}$ metrics, over the classical $p$-{\em Wasserstein} and {\em Bottleneck} metrics used between PDs. 
%%%%%%%%%%%%%%%%%%%%%%%%%%%%%
\subsection{Robustness to Topological Noise} \label{experiments_section_synthetic}

\begin{figure*}[htb!]
\centering
%% Shape 01
\subfloat{
\scalebox{0.7}{
\includegraphics[width = 0.1665\columnwidth]{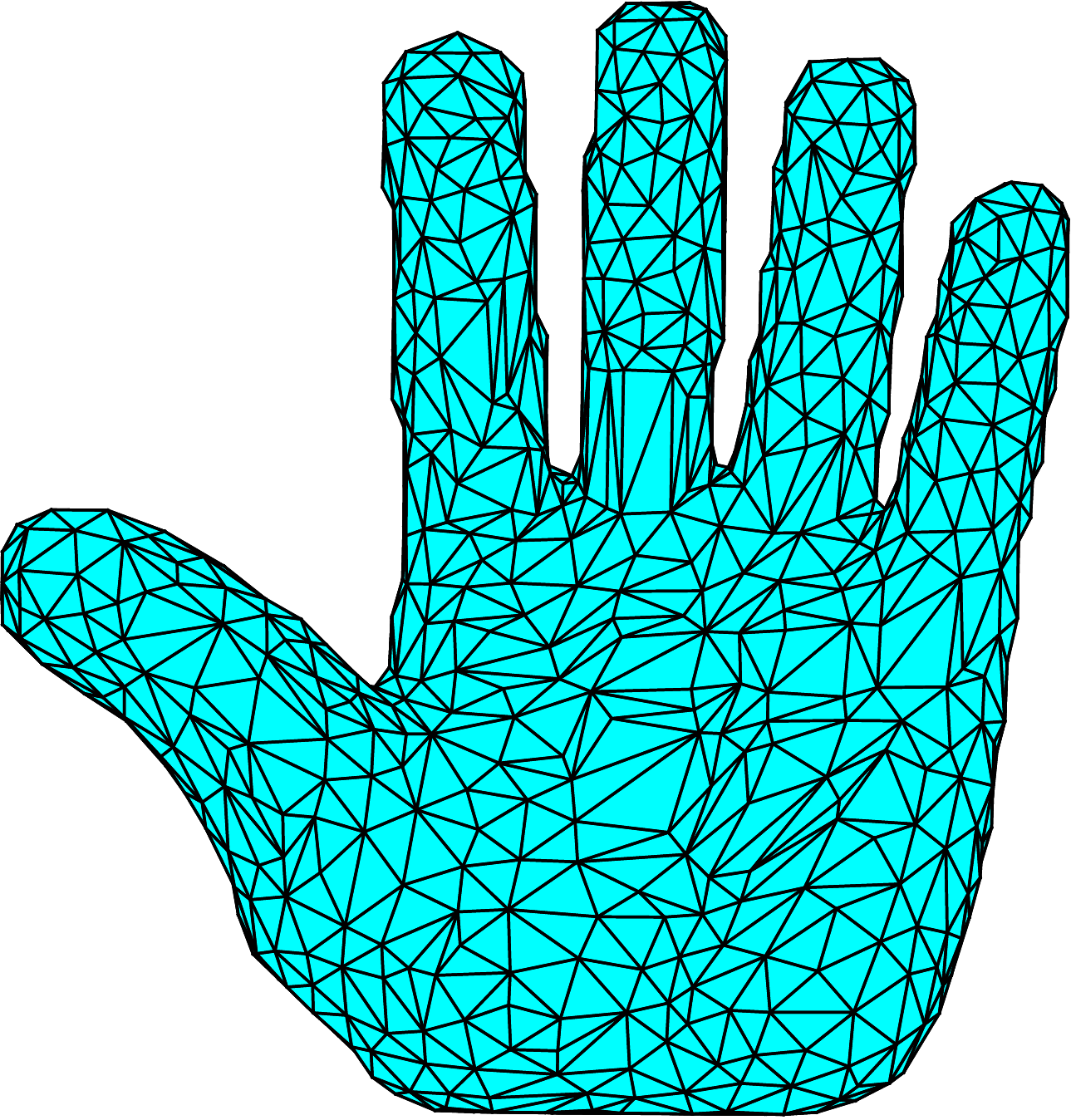}
\label{s1_3d}}}
\subfloat{
\scalebox{0.85}{
\includegraphics[width = 0.1665\columnwidth]{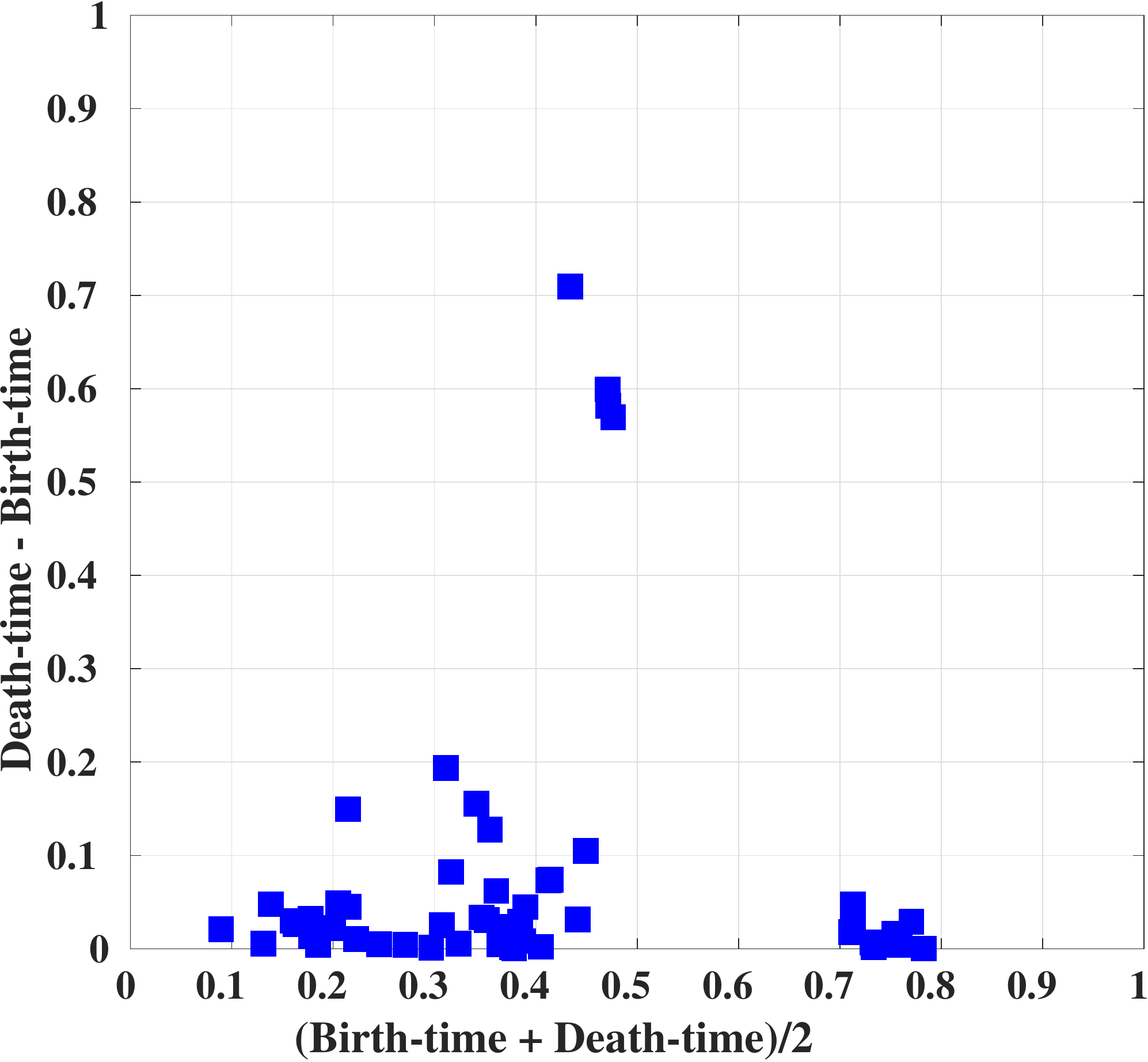}
\label{s1_pd}}}  
\subfloat{
\scalebox{1.025}{
\includegraphics[width = 0.1665\columnwidth]{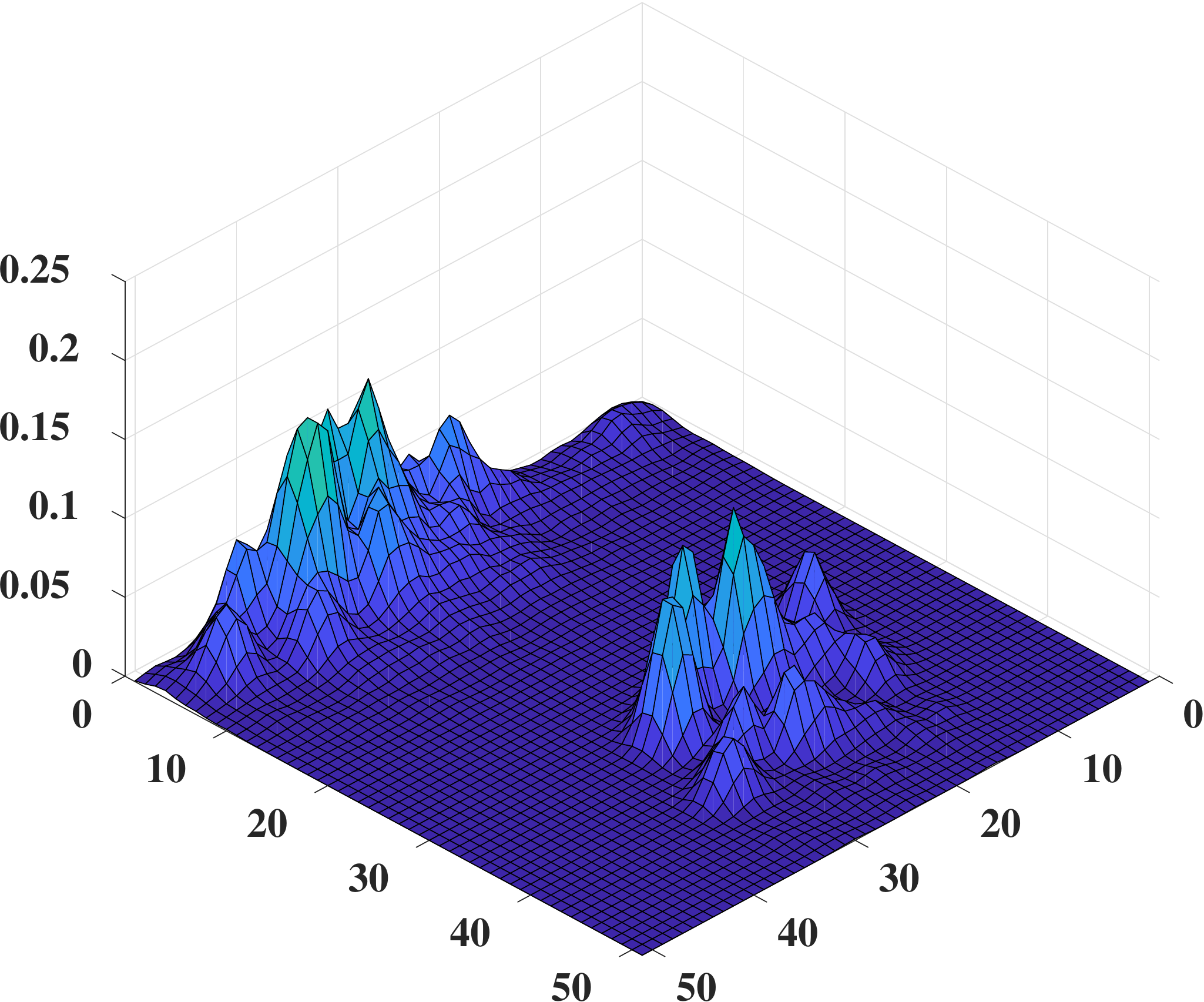}
\label{s1_gr}}}  
\subfloat{
\scalebox{1.025}{
\includegraphics[width = 0.1665\columnwidth]{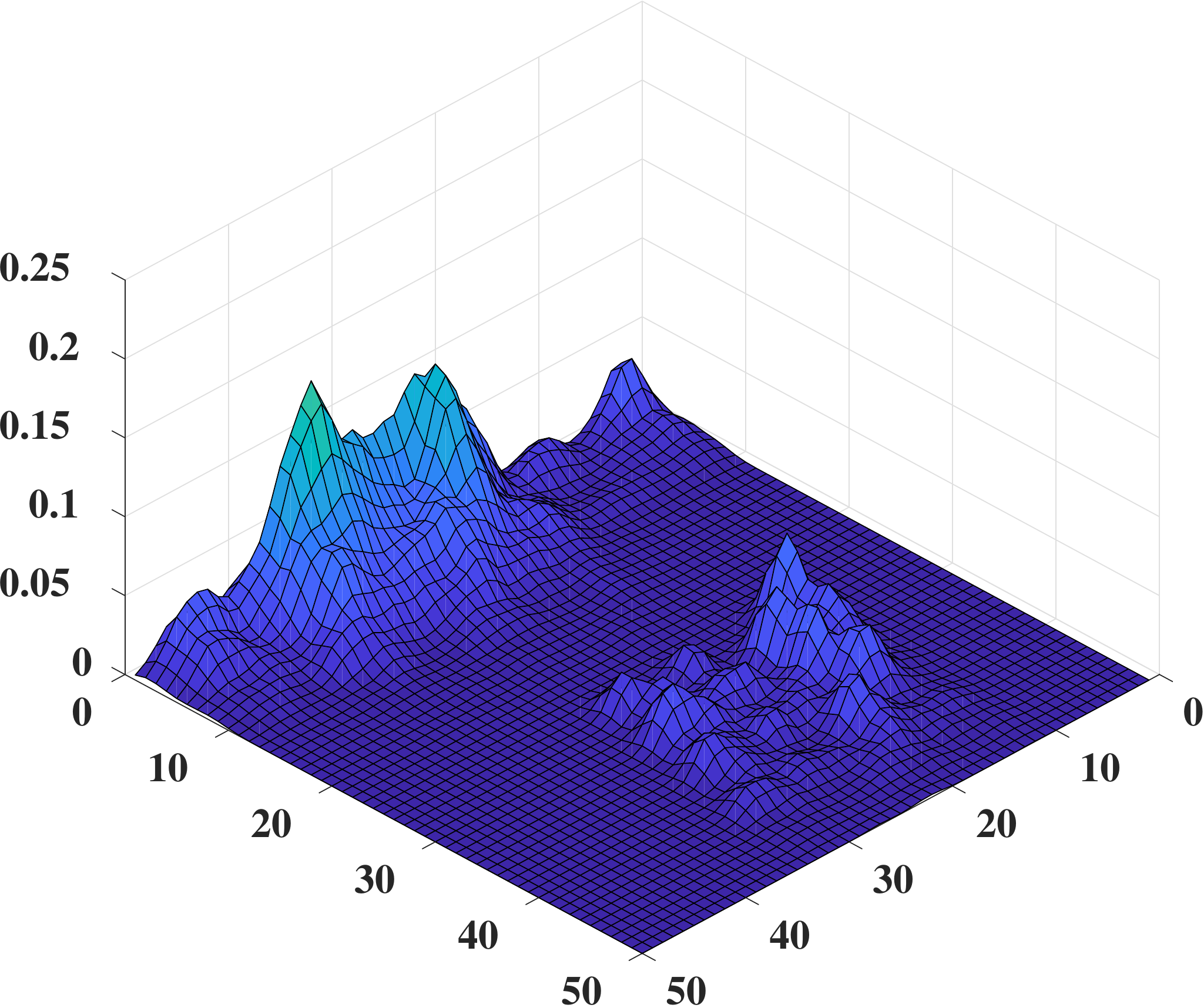}
\label{s11_gr}}} 
\subfloat{
\scalebox{0.85}{
\includegraphics[width = 0.1665\columnwidth]{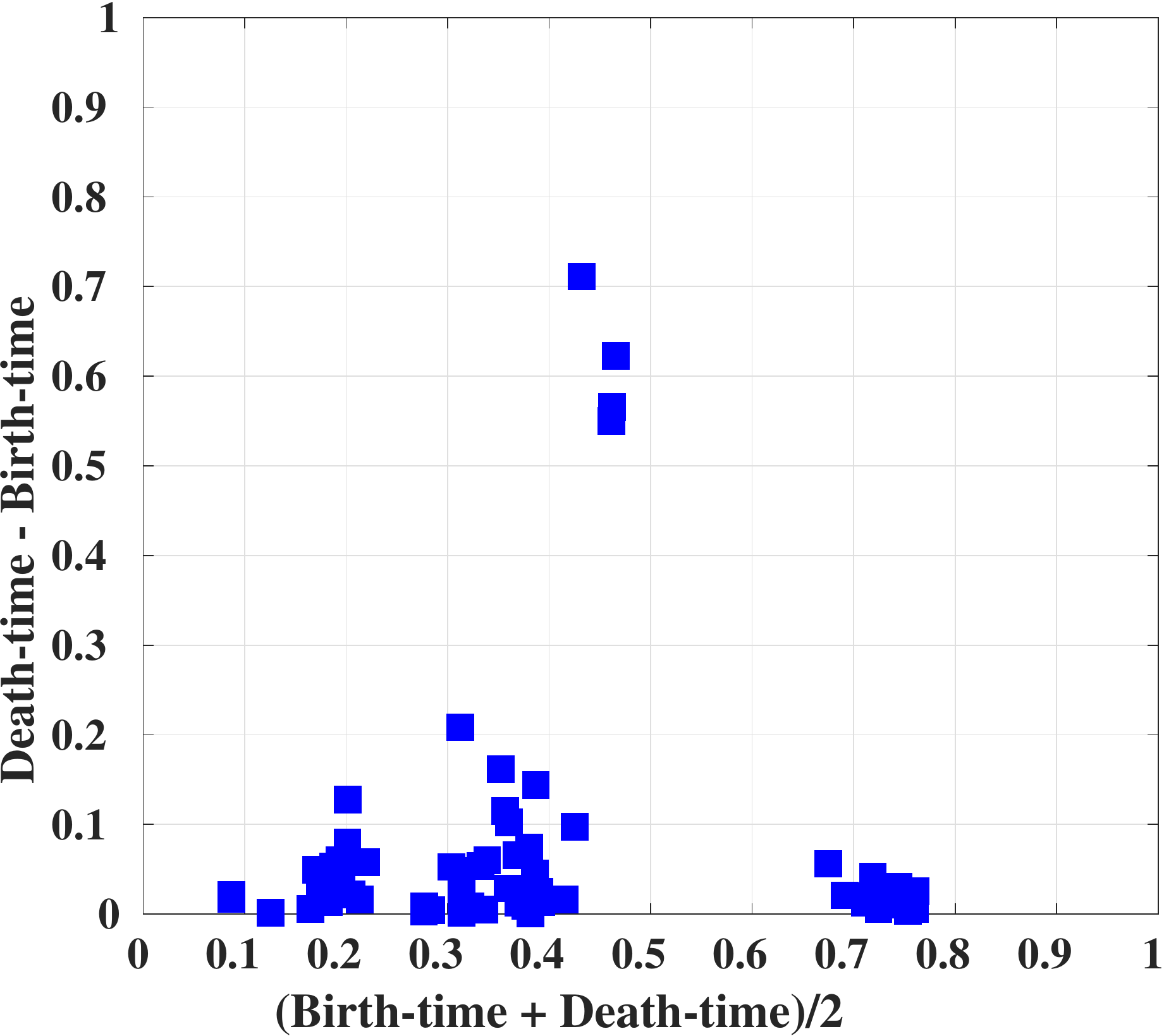}
\label{s11_pd}}} 
\subfloat{
\scalebox{0.7}{
\includegraphics[width = 0.1665\columnwidth]{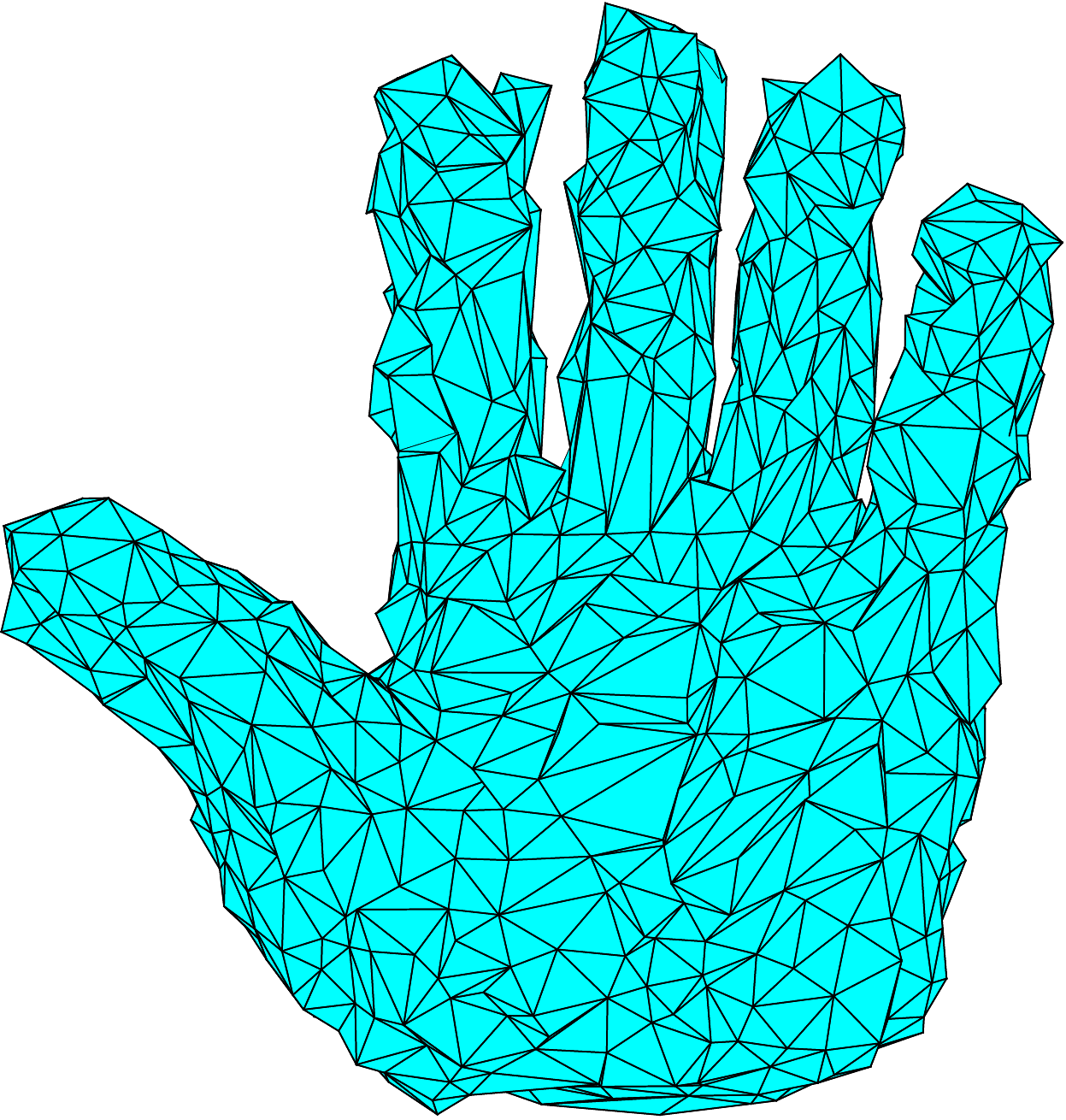}
\label{s11_3d}}}  

\vspace{-0.1 in}

%% Shape 02
\subfloat{
\scalebox{0.7}{
\includegraphics[width = 0.1665\columnwidth]{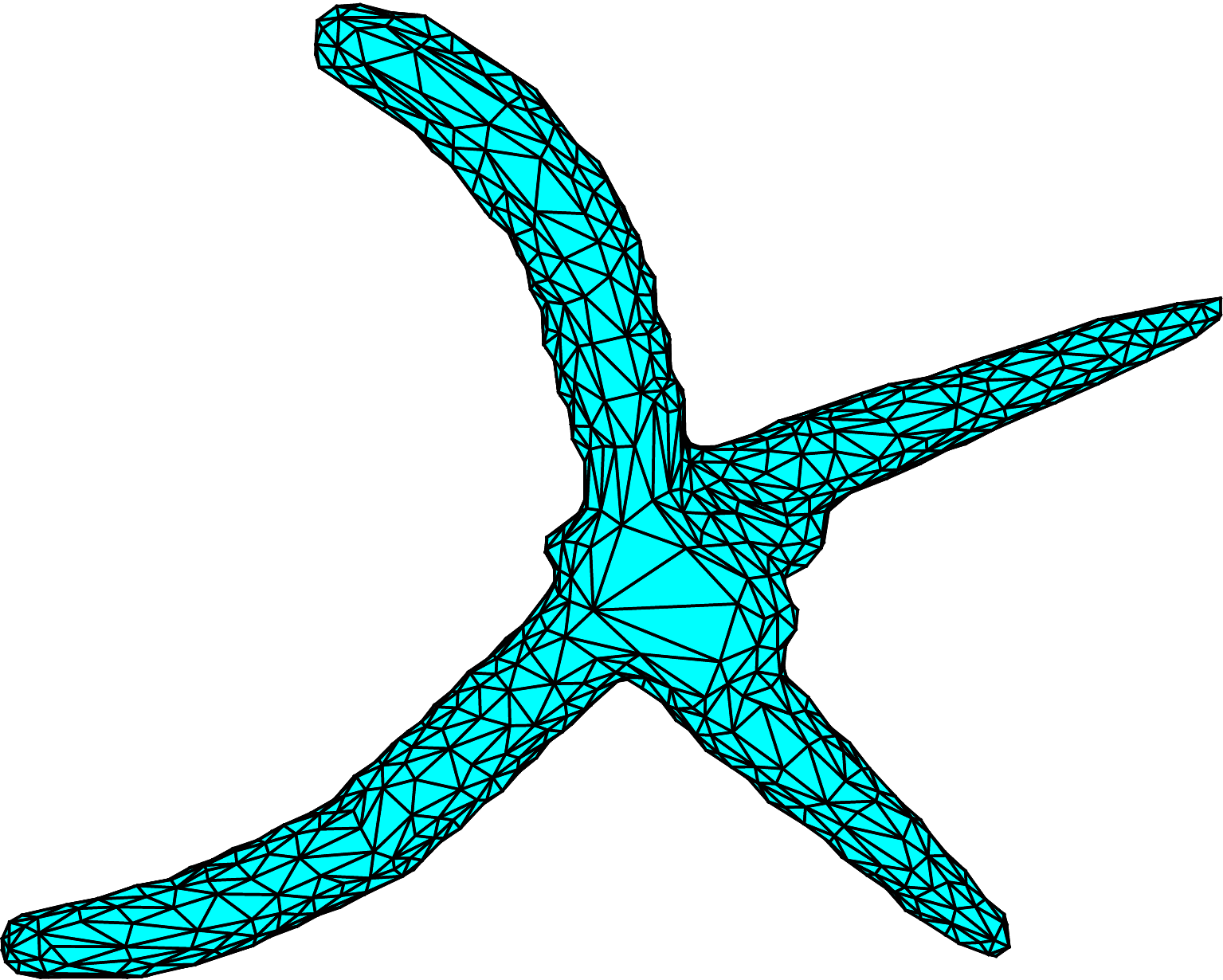}
\label{s2_3d}}}
\subfloat{
\scalebox{0.85}{
\includegraphics[width = 0.1665\columnwidth]{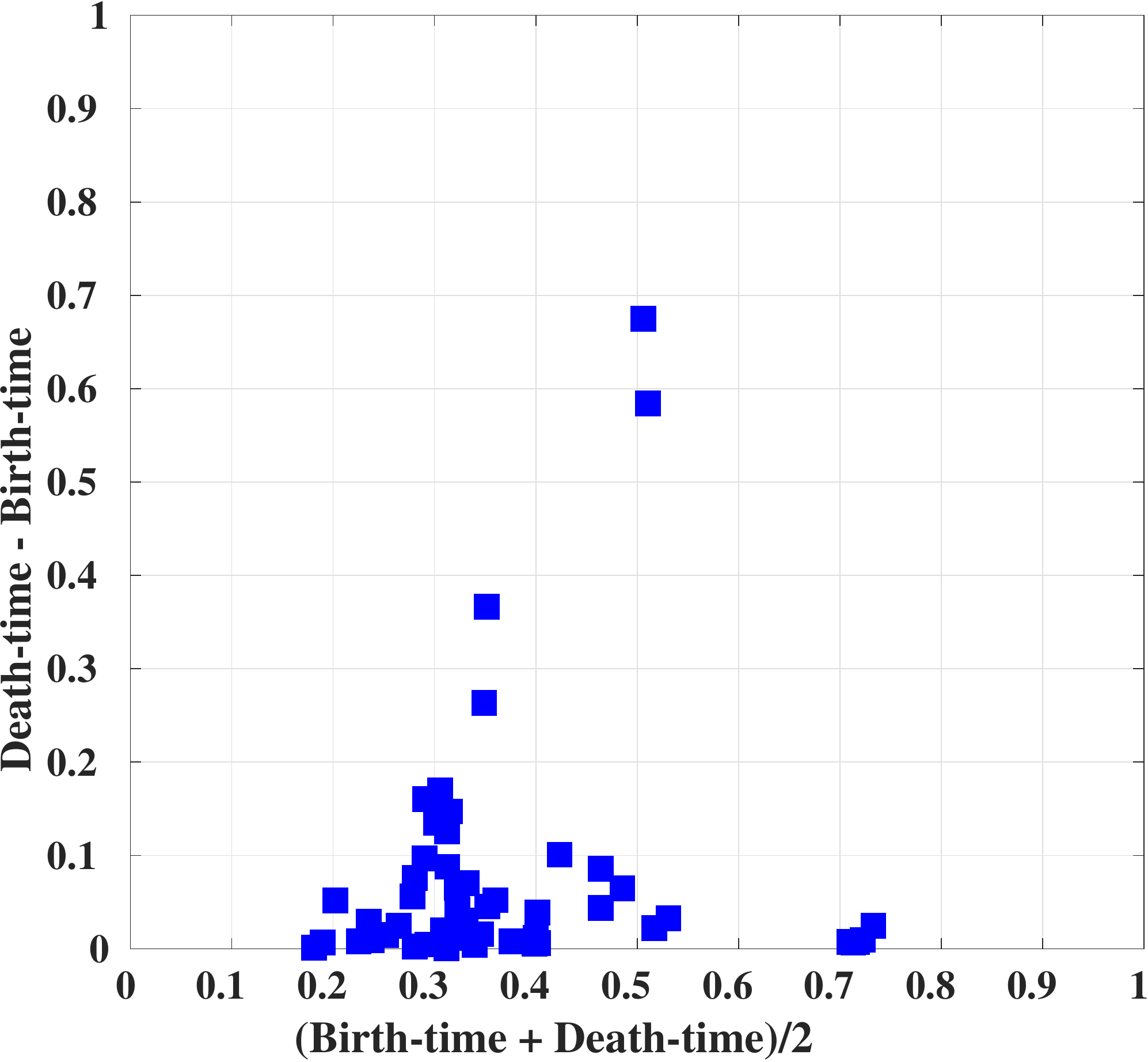}
\label{s2_pd}}}  
\subfloat{
\scalebox{1.025}{
\includegraphics[width = 0.1665\columnwidth]{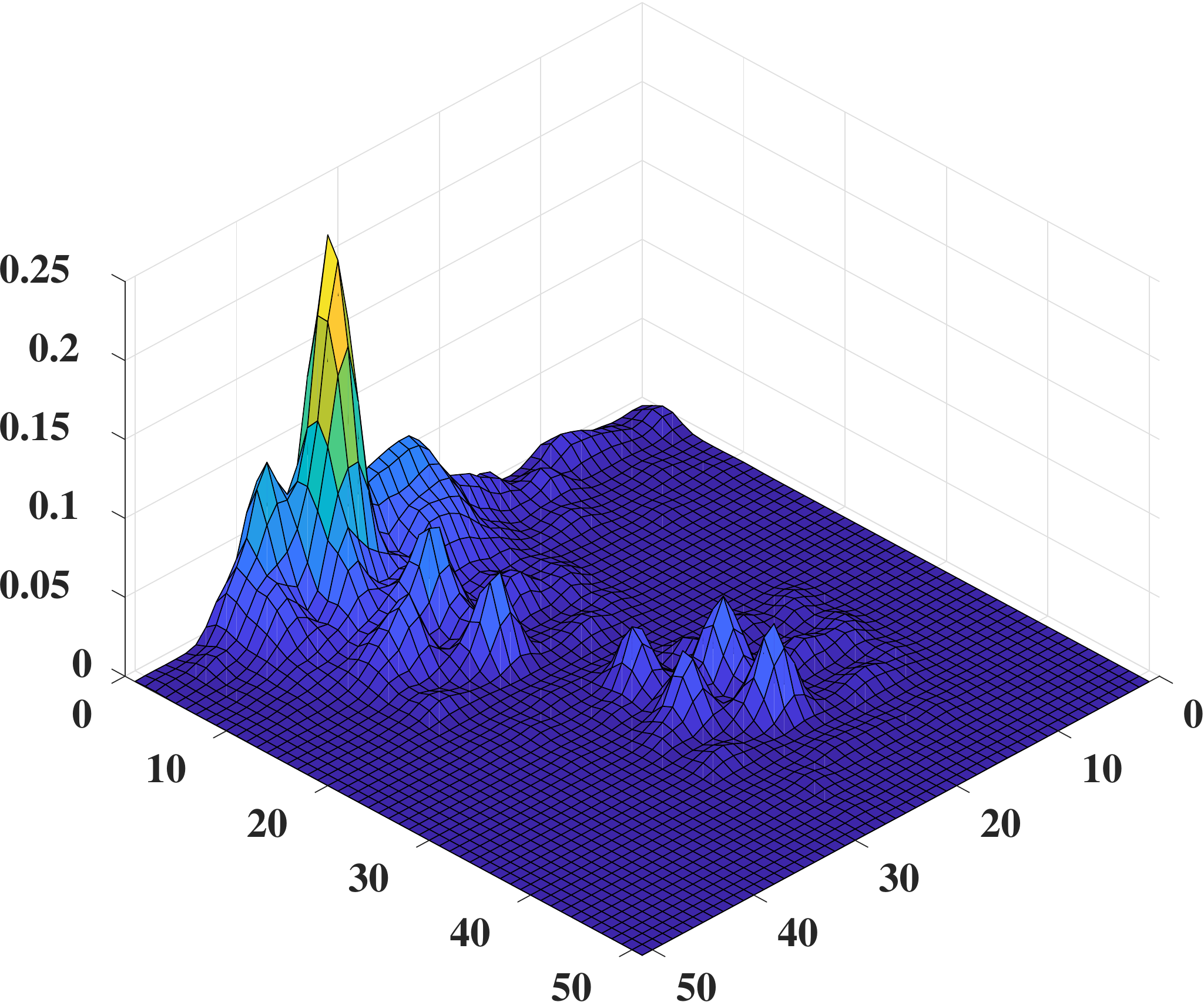}
\label{s2_gr}}}  
\subfloat{
\scalebox{1.025}{
\includegraphics[width = 0.1665\columnwidth]{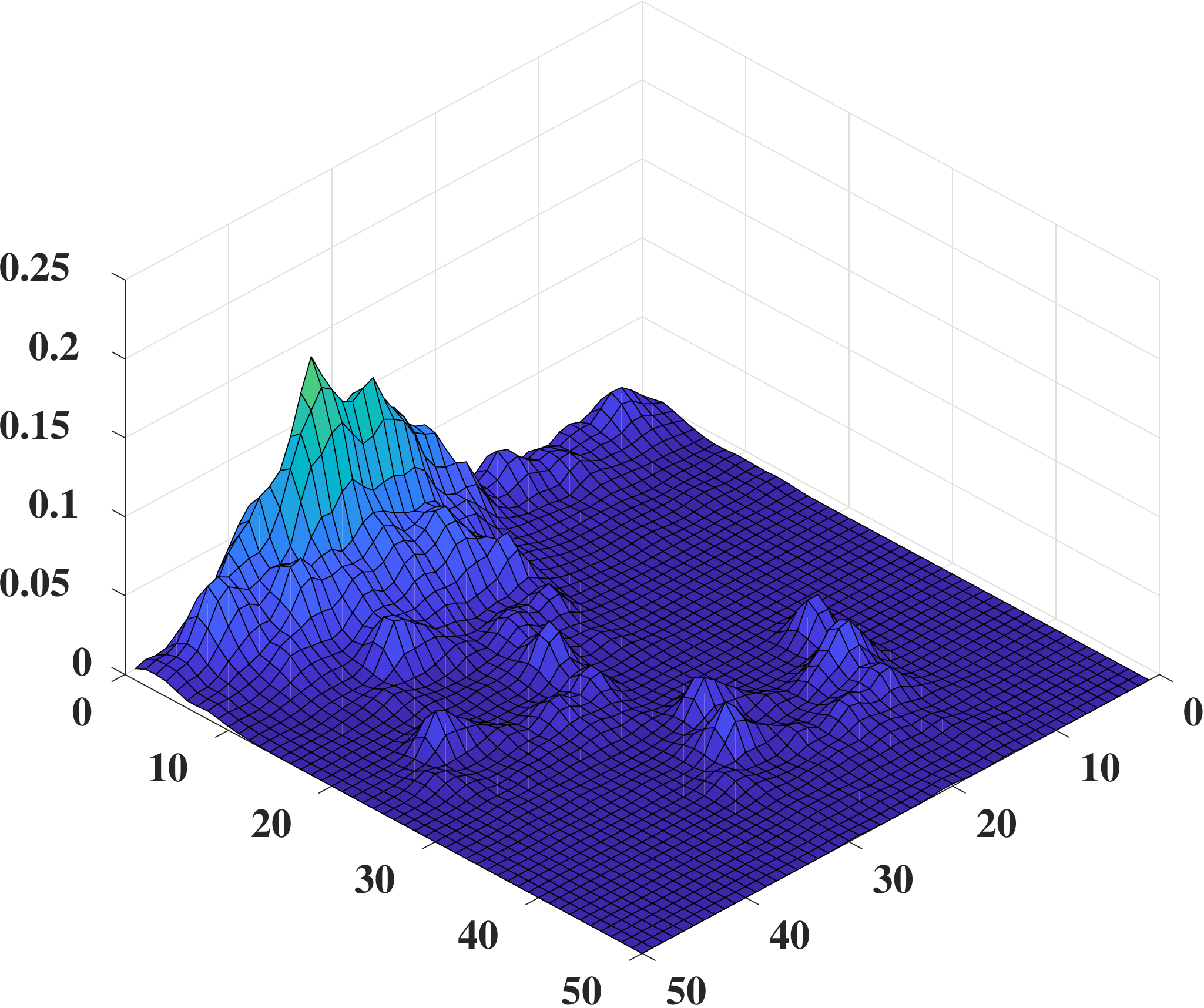}
\label{s21_gr}}} 
\subfloat{
\scalebox{0.85}{
\includegraphics[width = 0.1665\columnwidth]{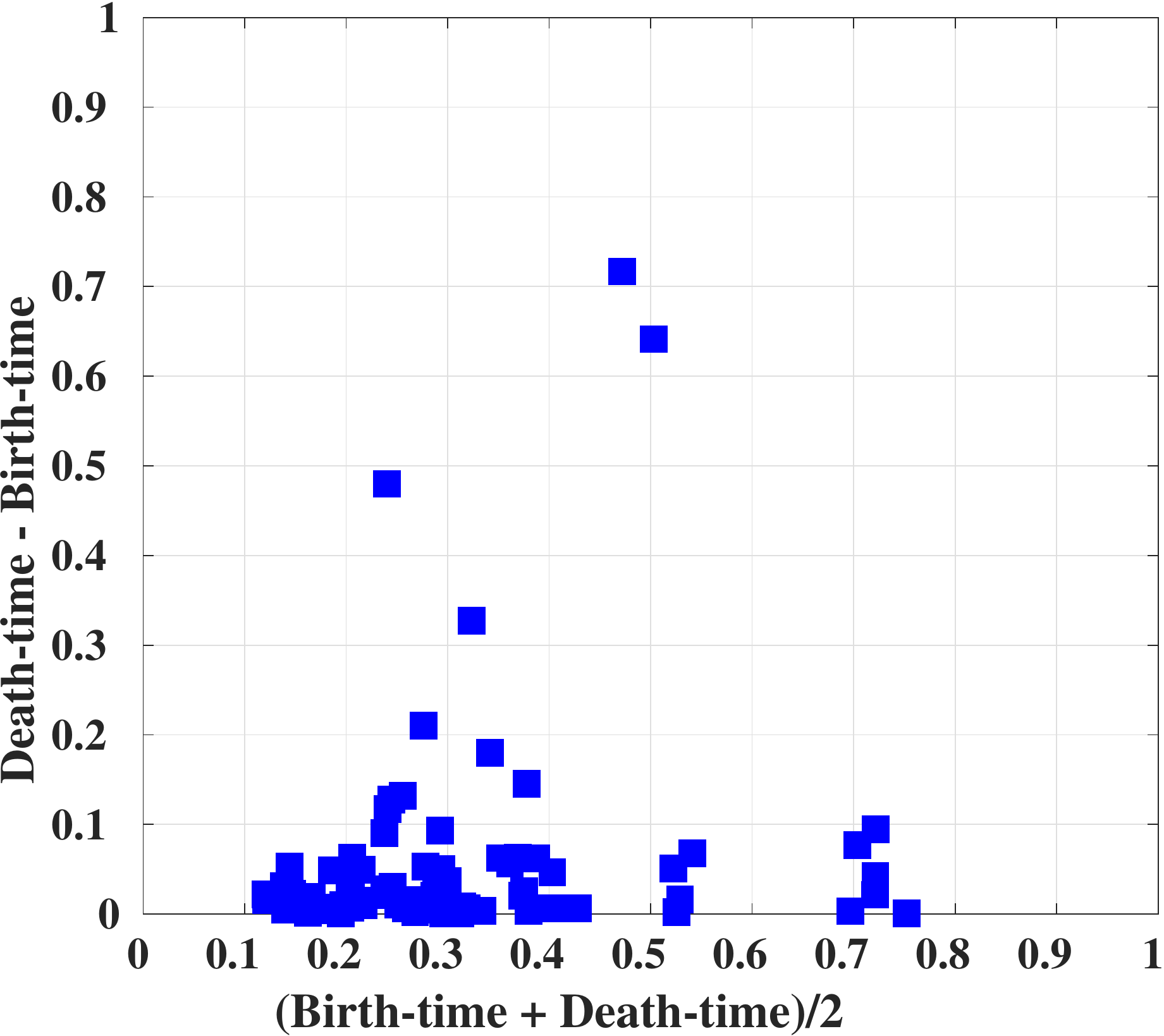}
\label{s21_pd}}} 
\subfloat{
\scalebox{0.7}{
\includegraphics[width = 0.1665\columnwidth]{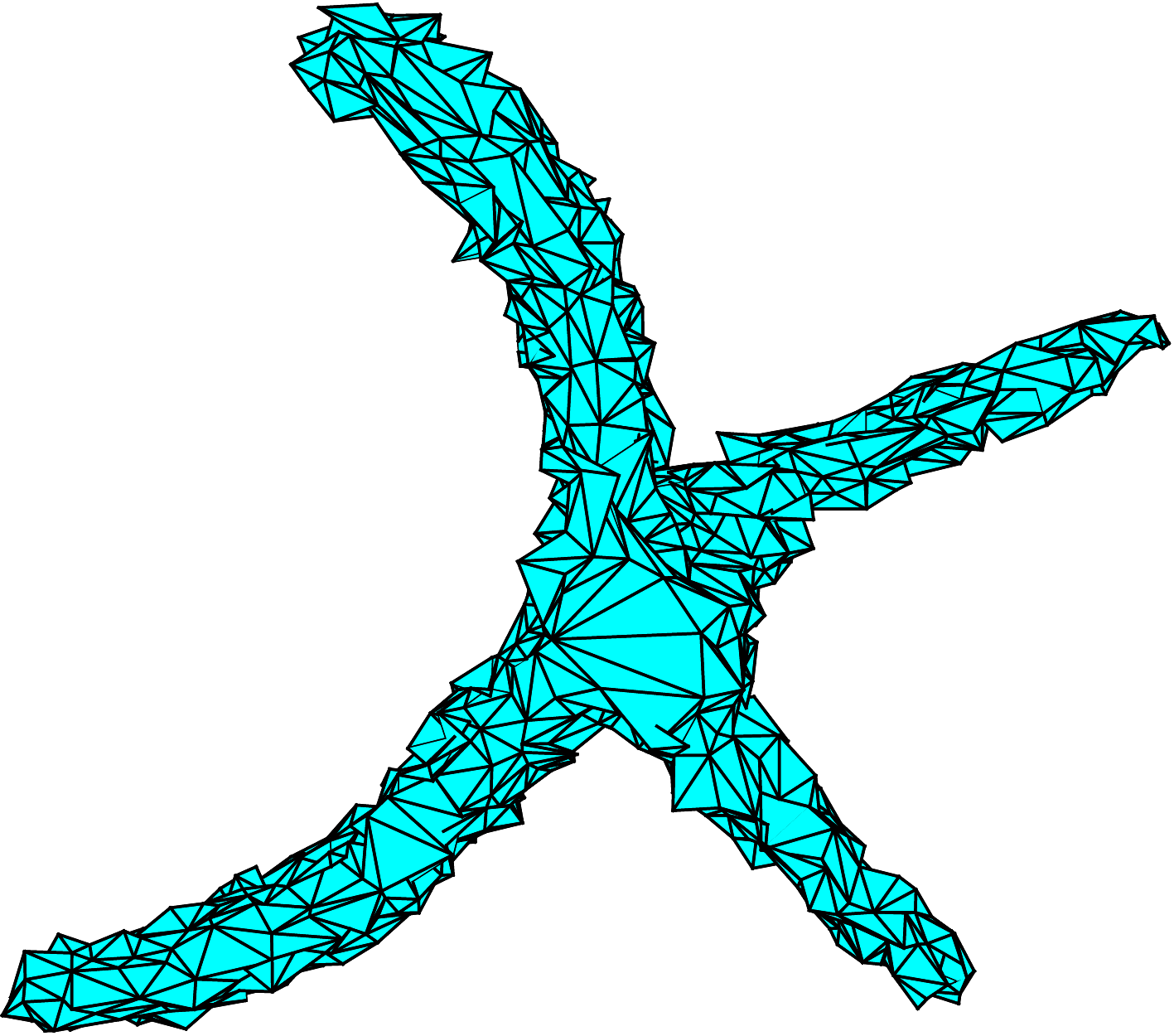}
\label{s21_3d}}}  

\vspace{-0.1 in}

%% Shape 03
\subfloat{
\scalebox{0.7}{
\includegraphics[width = 0.1665\columnwidth]{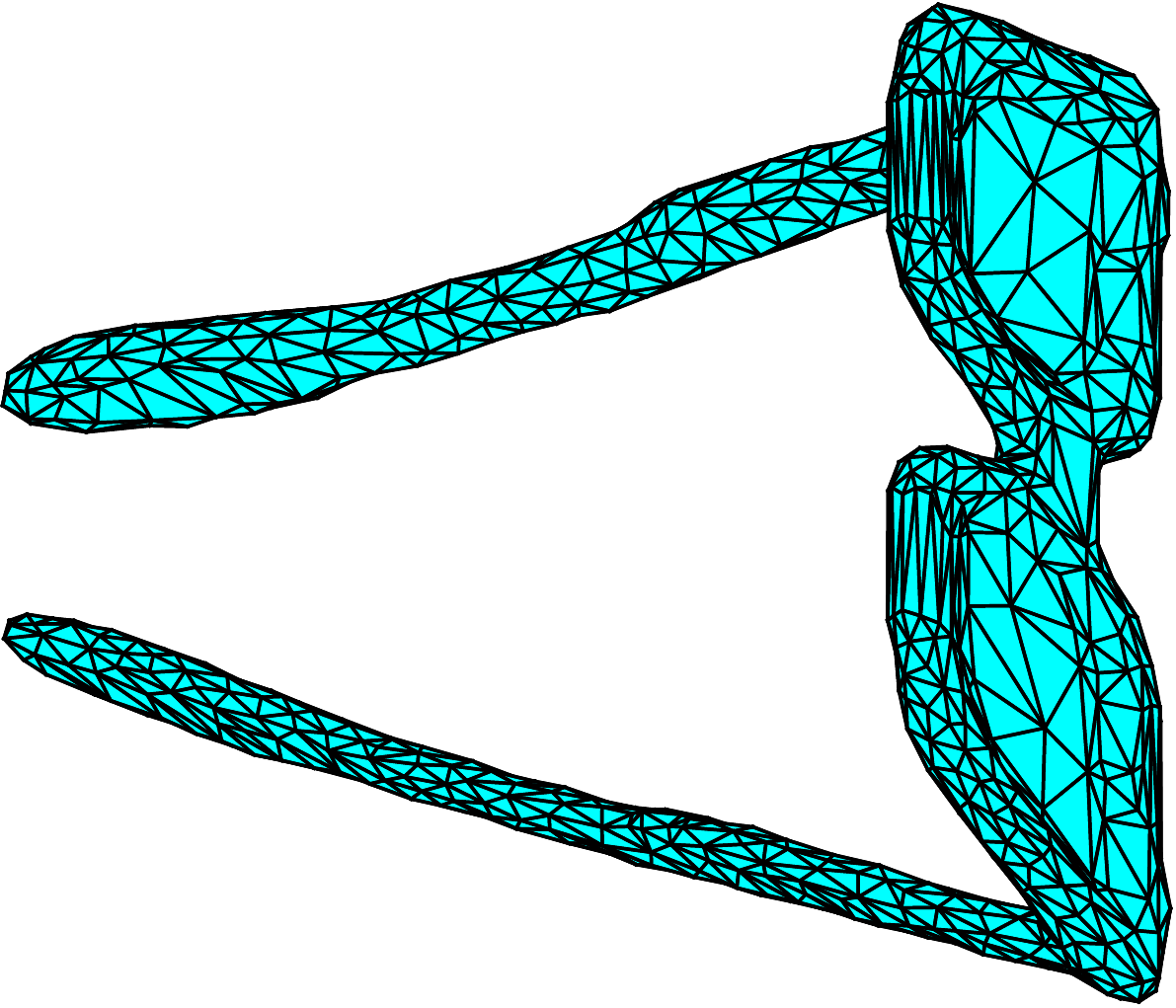}
\label{s3_3d}}}
\subfloat{
\scalebox{0.85}{
\includegraphics[width = 0.1665\columnwidth]{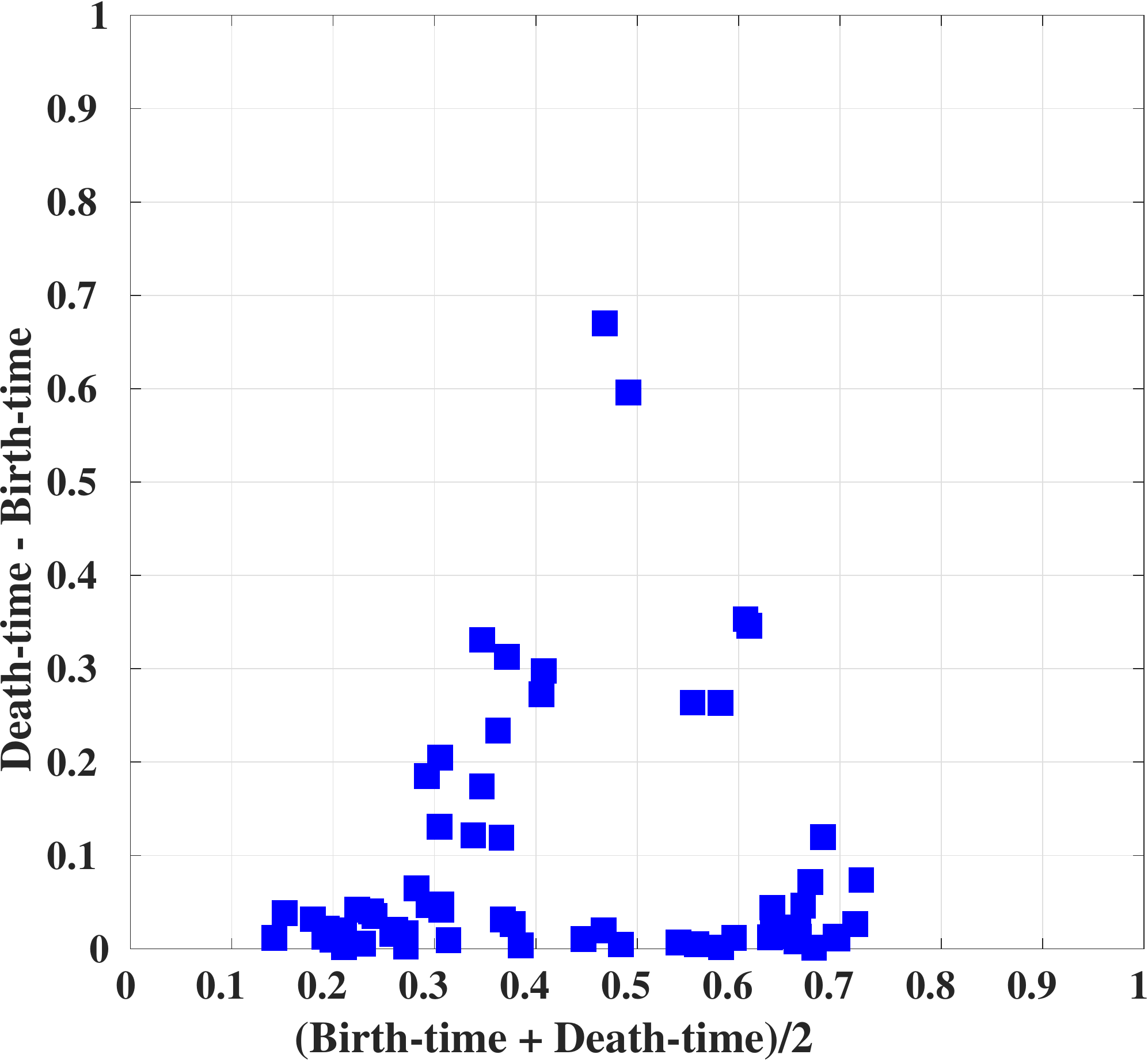}
\label{s3_pd}}}  
\subfloat{
\scalebox{1.025}{
\includegraphics[width = 0.1665\columnwidth]{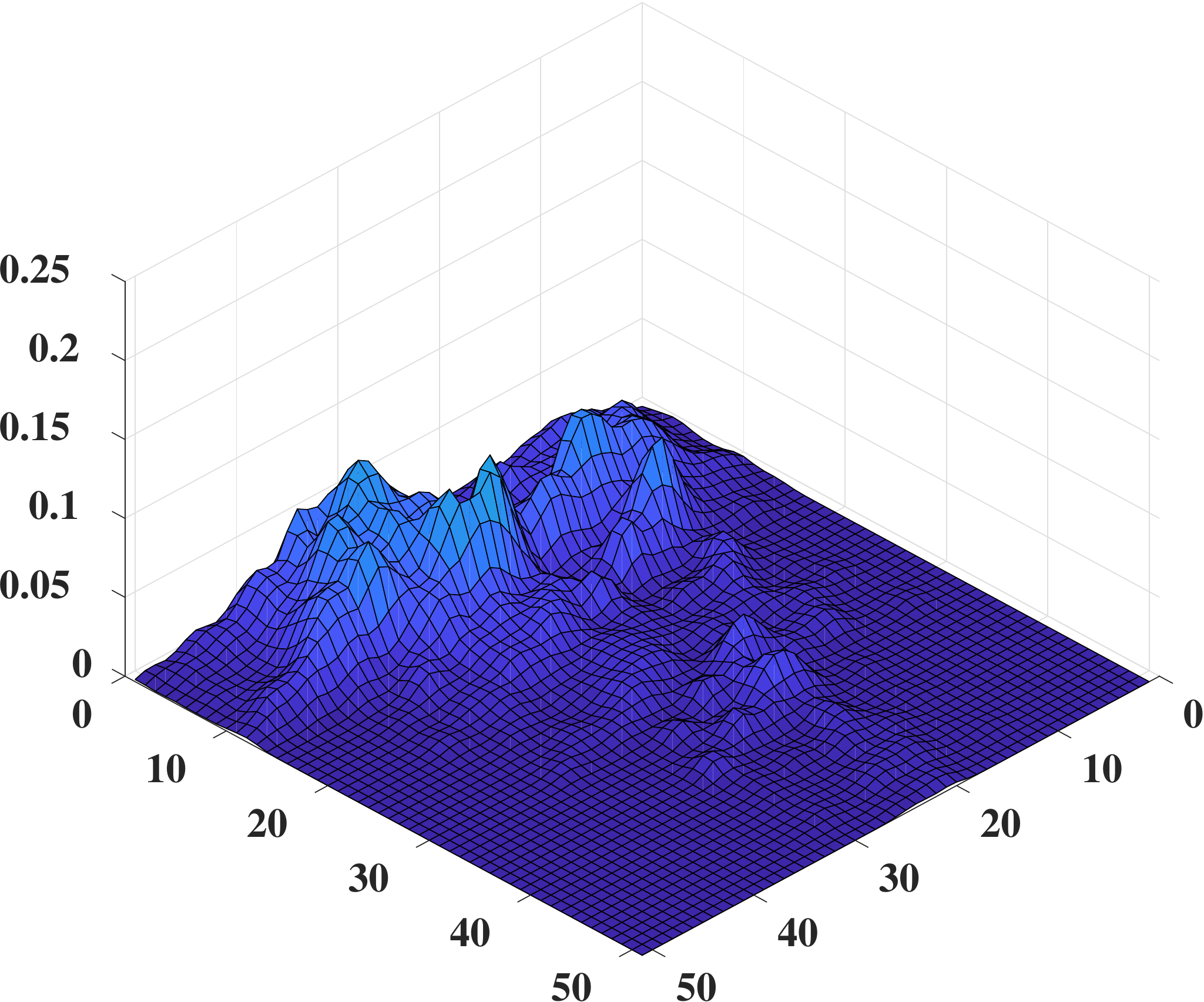}
\label{s3_gr}}}  
\subfloat{
\scalebox{1.025}{
\includegraphics[width = 0.1665\columnwidth]{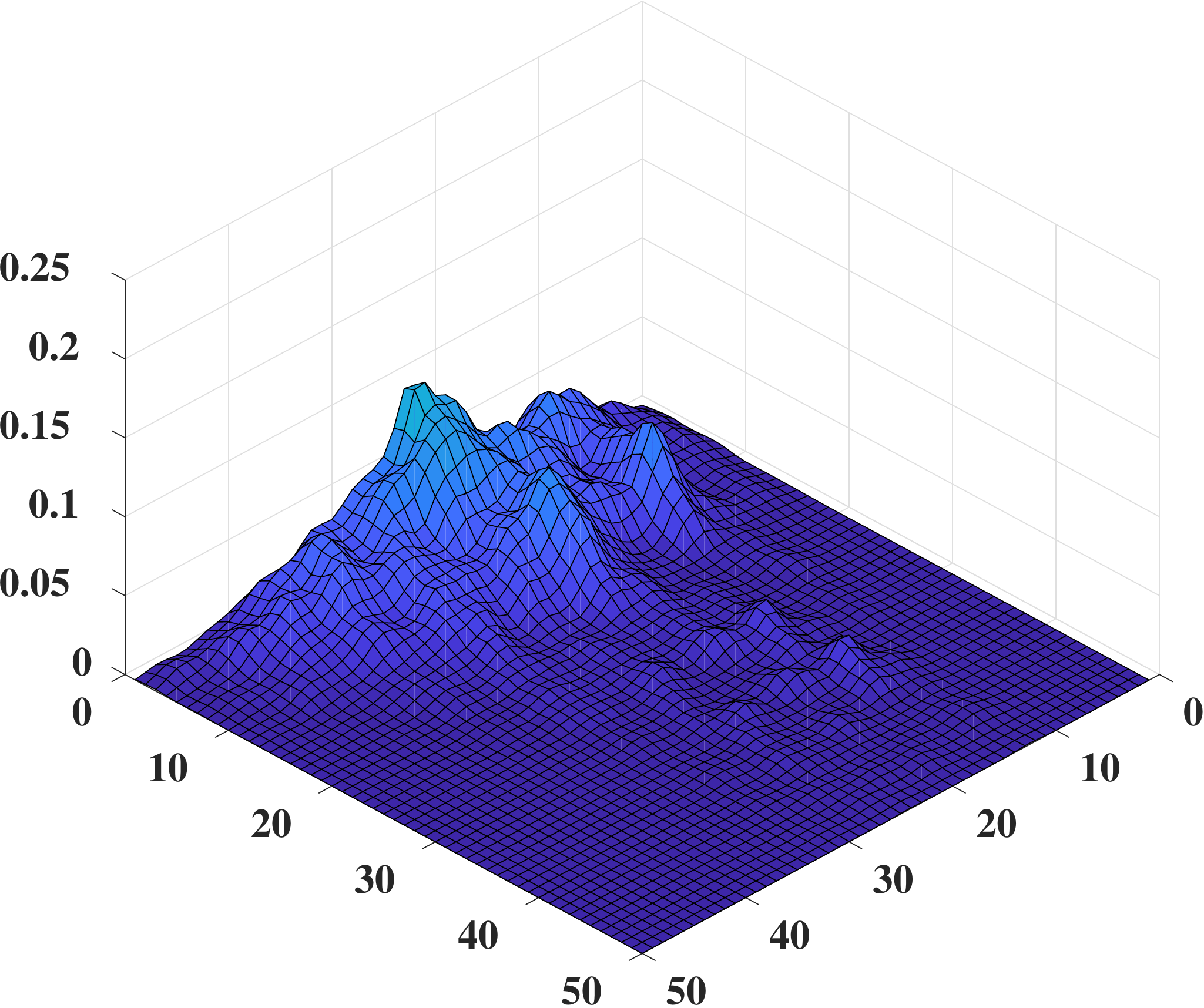}
\label{s31_gr}}} 
\subfloat{
\scalebox{0.85}{
\includegraphics[width = 0.1665\columnwidth]{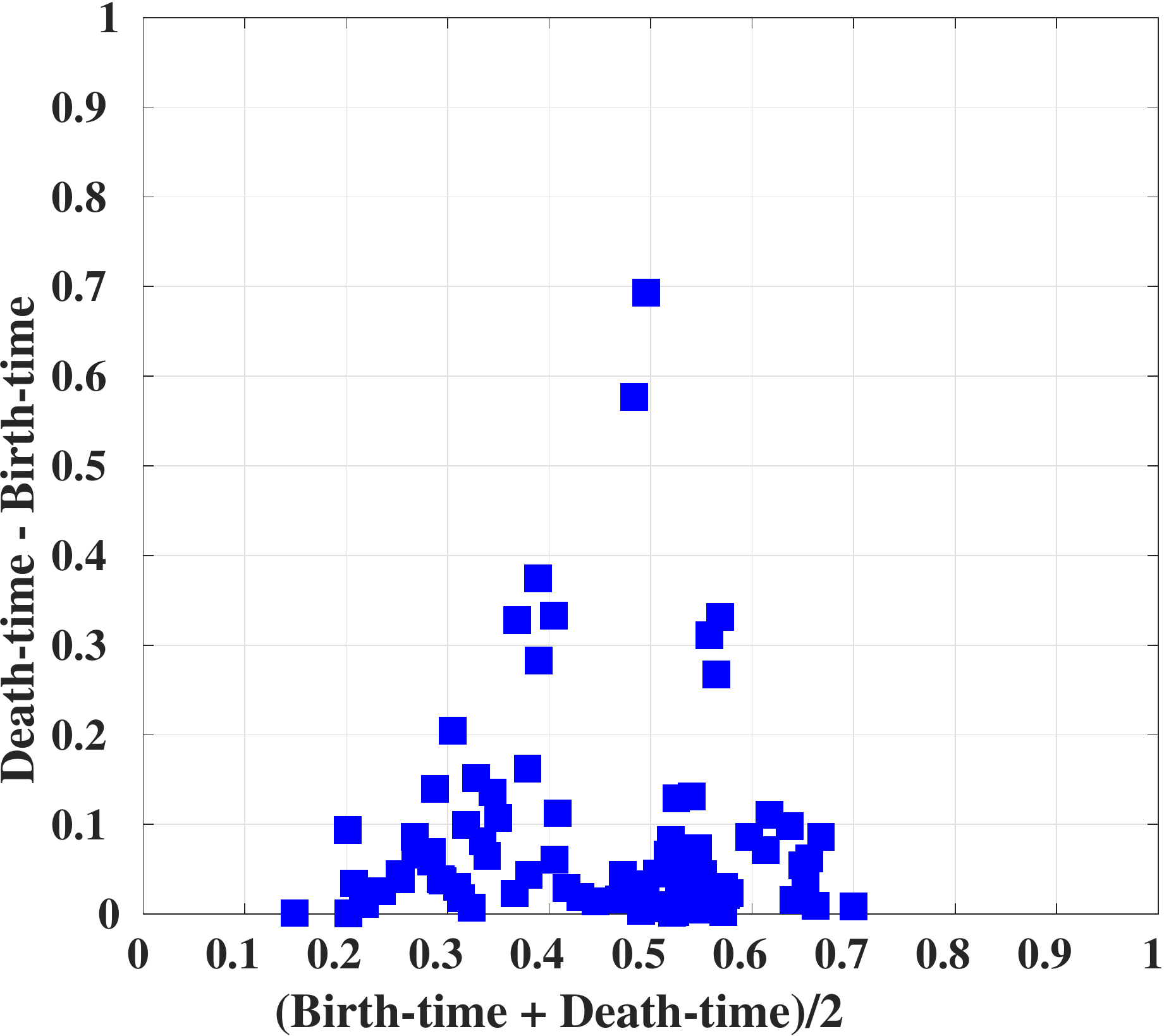}
\label{s31_pd}}} 
\subfloat{
\scalebox{0.7}{
\includegraphics[width = 0.1665\columnwidth]{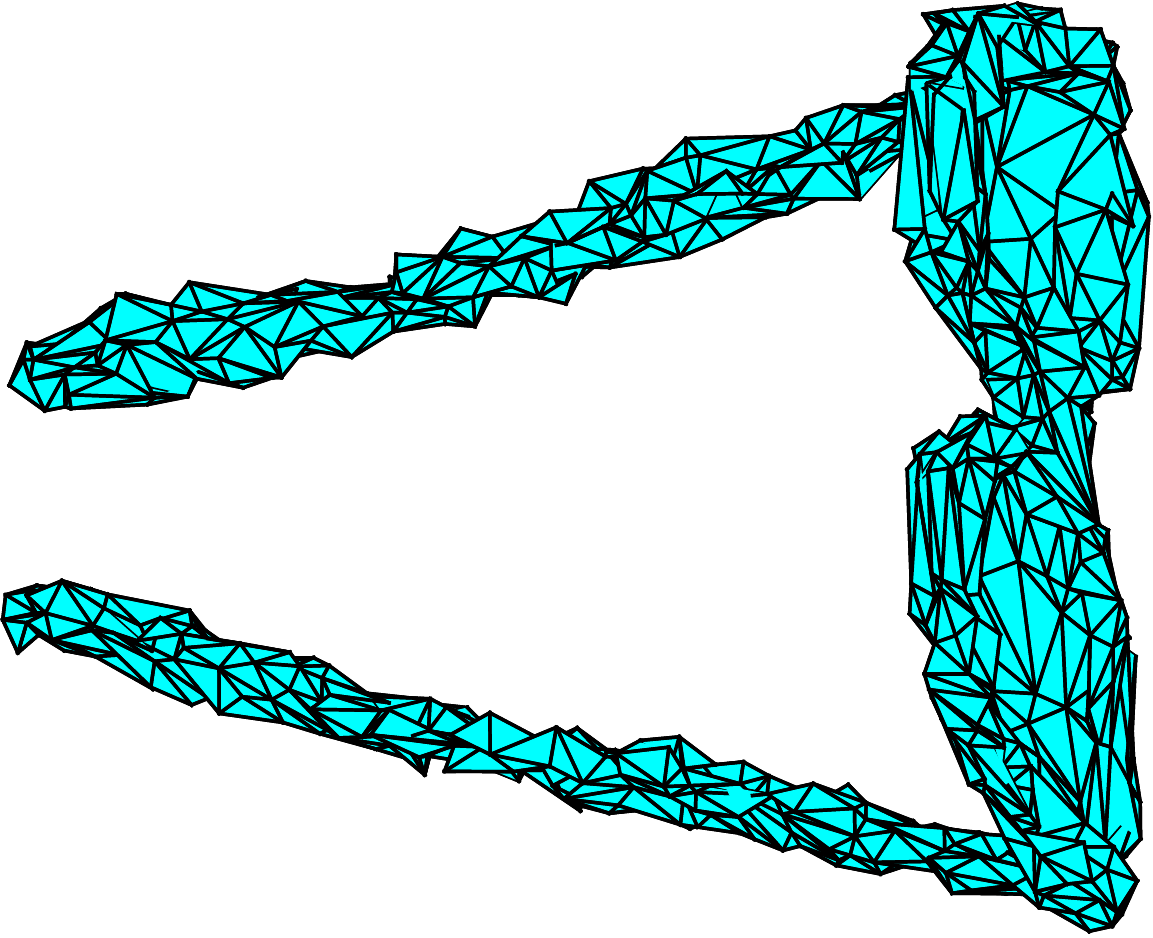}
\label{s31_3d}}}  

\vspace{-0.1 in}

%% Shape 05
\subfloat{
\scalebox{0.7}{
\includegraphics[width = 0.1665\columnwidth]{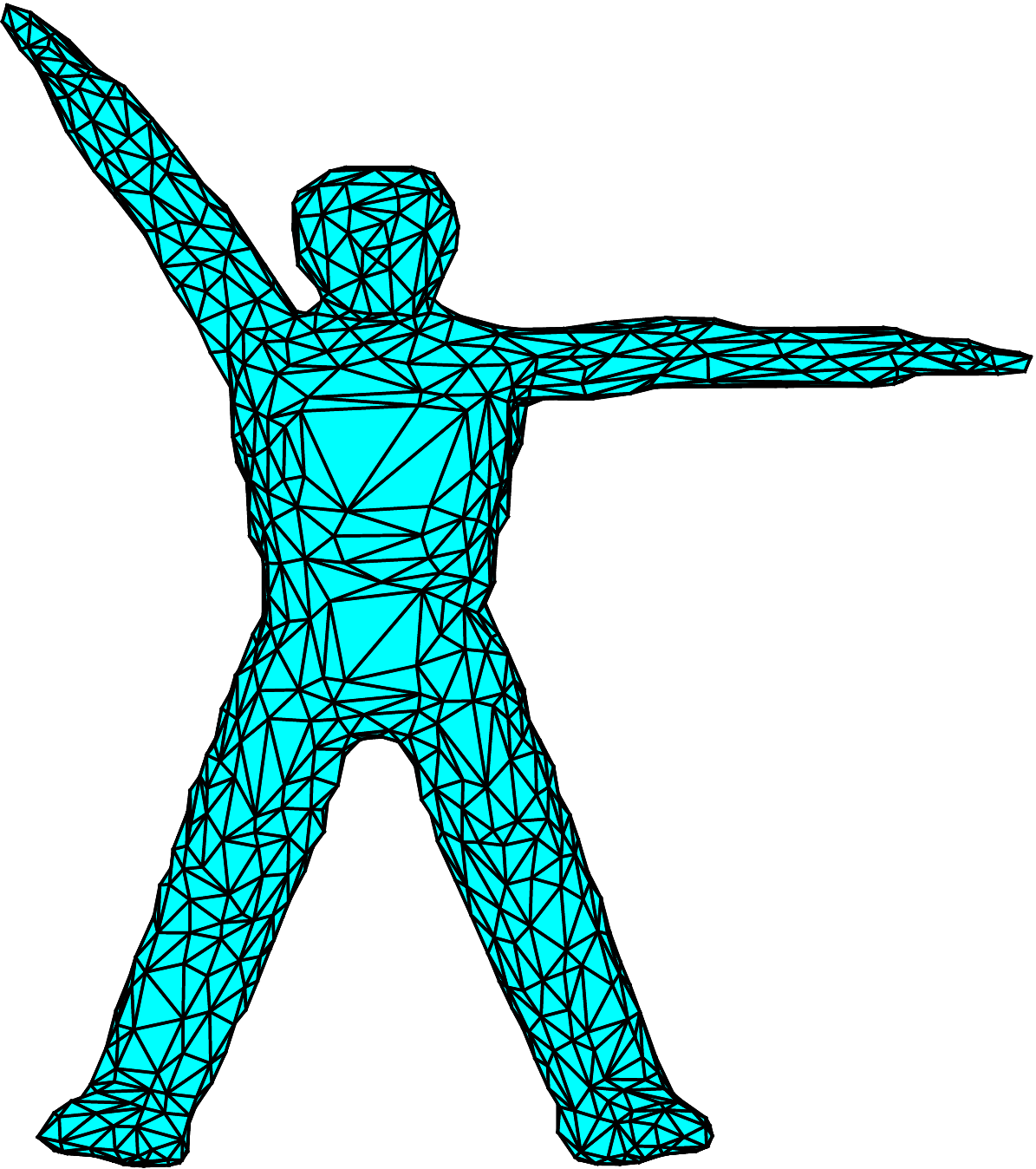}
\label{s5_3d}}}
\subfloat{
\scalebox{0.85}{
\includegraphics[width = 0.1665\columnwidth]{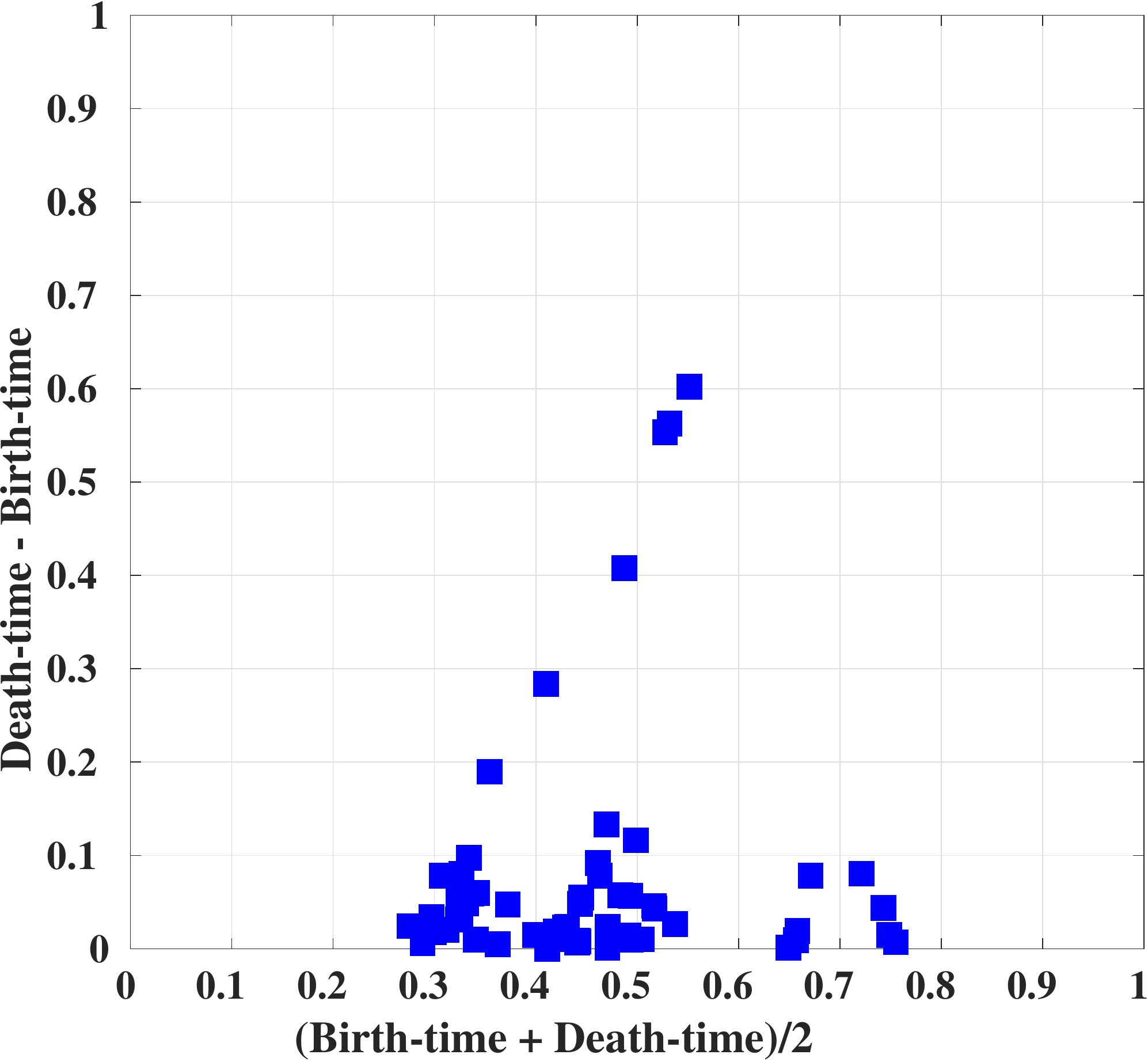}
\label{s5_pd}}}  
\subfloat{
\scalebox{1.025}{
\includegraphics[width = 0.1665\columnwidth]{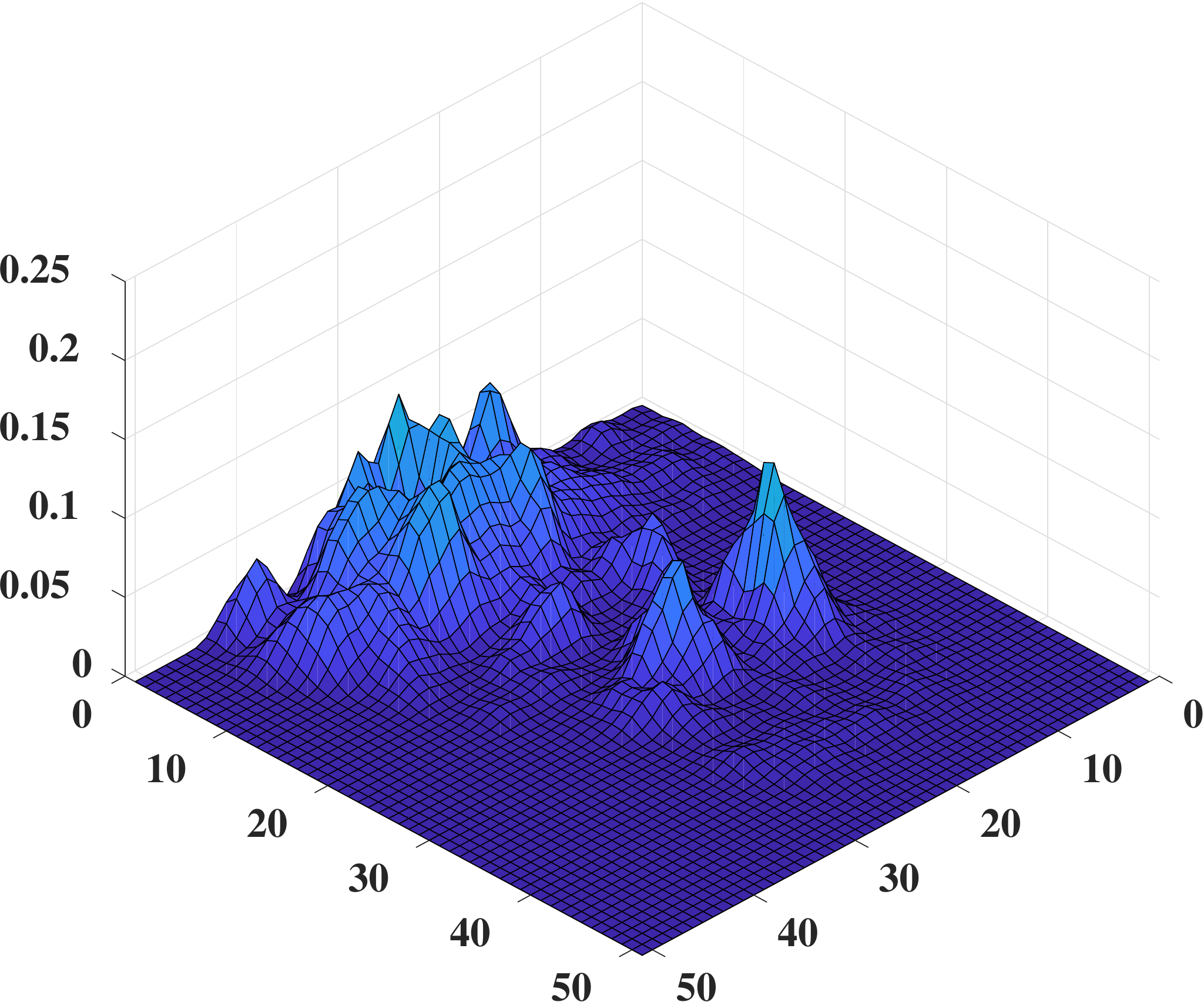}
\label{s5_gr}}}  
\subfloat{
\scalebox{1.025}{
\includegraphics[width = 0.1665\columnwidth]{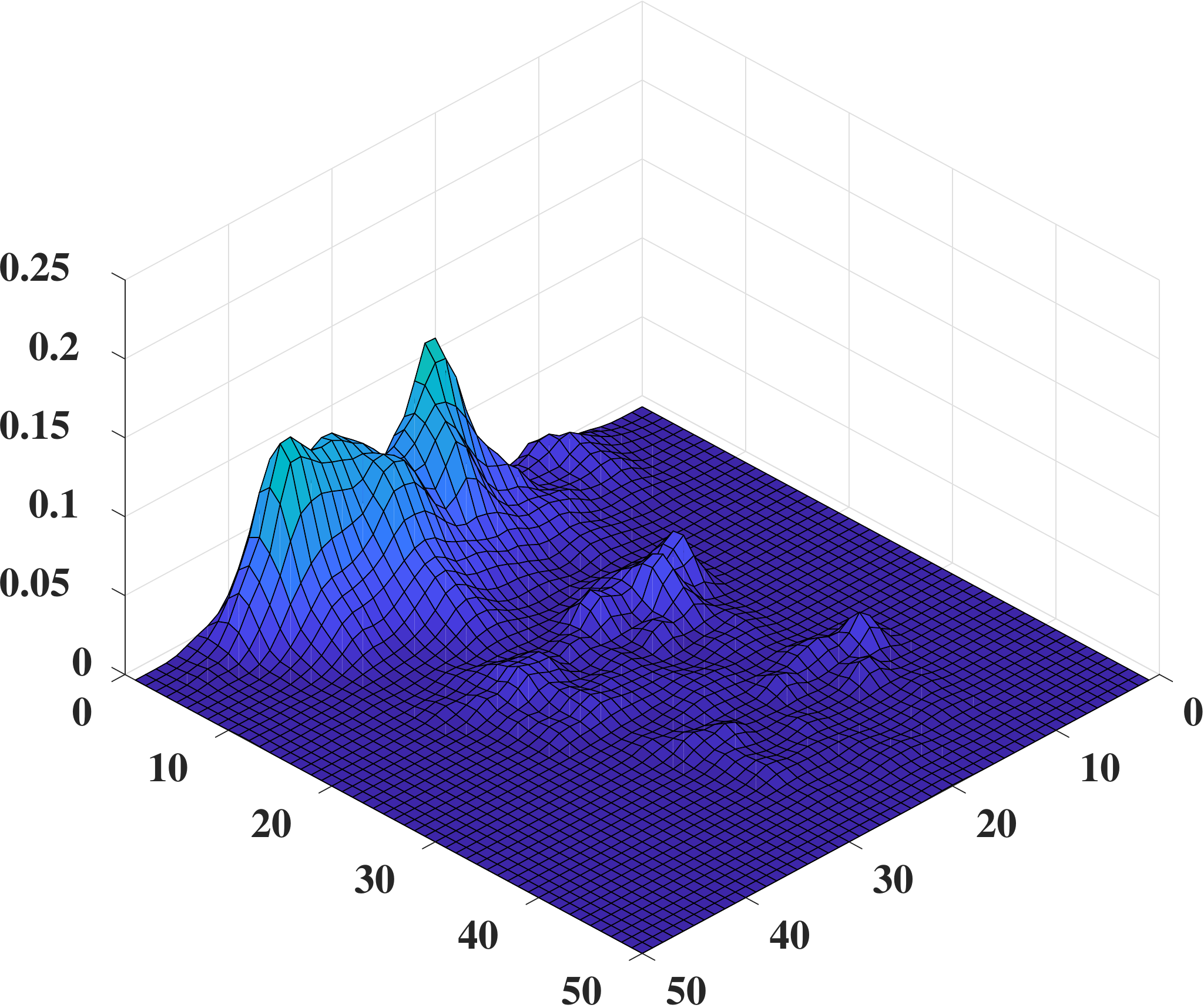}
\label{s51_gr}}} 
\subfloat{
\scalebox{0.85}{
\includegraphics[width = 0.1665\columnwidth]{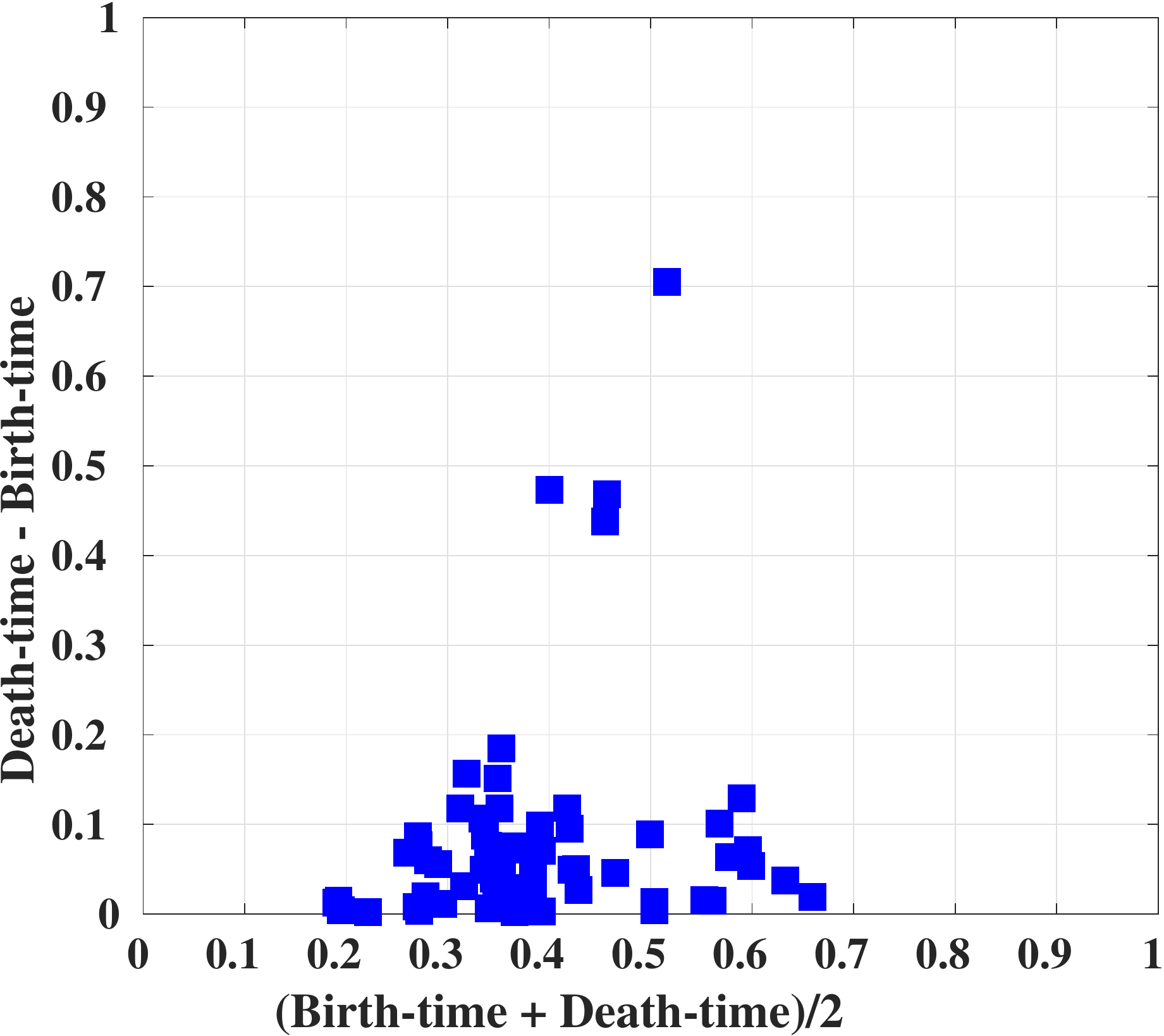}
\label{s51_pd}}} 
\subfloat{
\scalebox{0.7}{
\includegraphics[width = 0.1665\columnwidth]{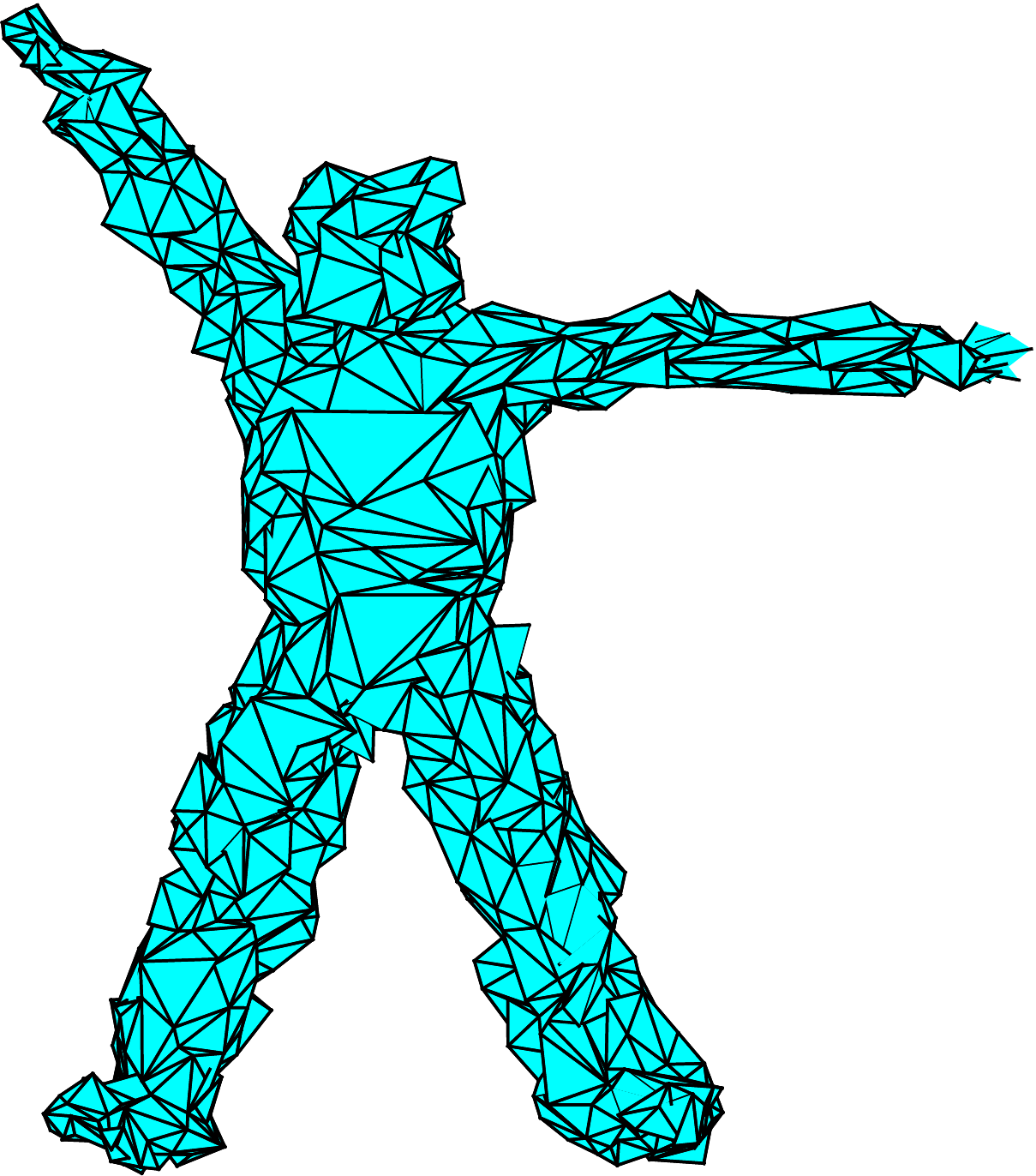}
\label{s51_3d}}}  

\vspace{-0.075 in}

\caption{Illustration of PD and PTS representations for 4 shapes and their noisy variants. Columns 1 and 6 represent the 3D shape with triangular mesh faces; columns 2 and 5 show the corresponding  9$^\text{th}$ dimension SIHKS function-based PDs. columns 3 and 4 depict the PTS feature of the PD for the original and noisy shapes respectively. A zero mean Gaussian noise with standard deviation 1.0 is applied on the original shapes in column 1 to get the corresponding noisy variant in column 6. The PTS representation shown is the largest left singular vector (reshaped to a 2D matrix) obtained after applying SVD on the set of 2D PDFs and lies on the $\mathbb{G}_{1,n}$ space.}
\label{Synthetic_experiment_fig}
\end{figure*}

%(shown in figure \ref{syn_exp_shapes})

We conduct this experiment on $10$ randomly chosen shapes from the SHREC 2010 dataset \cite{lian2010shrec}. The dataset consists of 200 near-isometric watertight 3D shapes with articulating parts, equally divided into 10 classes. Each 3D mesh is simplified to 2000 faces. The 10 shapes used in the experiment  are denoted as $\mathcal{S}_i$, $i=1,2, \dots,10$. The minimum bounding sphere for each of these shapes has a mean radius of 54.4 with standard deviation of 3.7 centered at $(64.4, 63.4, 66.0)$ with coordinate-wise standard deviations of $(3.9, 4.1, 4.9)$ respectively. Next, we generate 100 sets of shapes, infused with topological noise. Topological noise is applied by changing the position of the vertices of the triangular mesh face, which results in changing its normal. We do this by applying a zero-mean Gaussian noise to the vertices of the original shape, with the standard deviation $\sigma$ varied from 0.1 to 1 in steps of 0.1. For each shape $\mathcal{S}_i$, its 10 noisy shapes with different levels of topological noise are denoted by $\mathcal{N}_{i,1}, \dots, \mathcal{N}_{i,10}$.

\setlength{\intextsep}{5pt}%
\begin{table*}[b!]
  \centering
    \scalebox{0.675}{
    \begin{tabular}{|c|c|c|c|c|c|c|c|c|c|c|c|c|}
      \hline
      \textbf{Method} & $\bm{\mathcal{N}_{i,1}}$ & $\bm{\mathcal{N}_{i,2}}$ & $\bm{\mathcal{N}_{i,3}}$ & $\bm{\mathcal{N}_{i,4}}$ & $\bm{\mathcal{N}_{i,5}}$ & $\bm{\mathcal{N}_{i,6}}$ & $\bm{\mathcal{N}_{i,7}}$ & $\bm{\mathcal{N}_{i,8}}$ & $\bm{\mathcal{N}_{i,9}}$ & $\bm{\mathcal{N}_{{i,10}}}$  & \makecell{\textbf{Average} \\ \textbf{Accuracy (\%)}} & \makecell{\textbf{Average Time} \\ \textbf{Taken \textbf{(}$\bm{10^{-4}}$ \textbf{sec)}}} \\
      \hline
      {PD ($1$-{\em Wasserstein})} & 100.00 & 100.00 & 100.00 & 99.90 & 100.00 & 99.80 & 99.60 & 99.00 & 96.60 & 94.40 & 98.93 & 256.00\\
      \hline
      {PD ($2$-{\em Wasserstein})} & 97.50 & 98.00 & 98.10 & 97.20 & 97.20 & 96.00 & 94.40 & 92.80 & 90.30 & 88.50 & 95.00 & 450.00\\
      \hline
      {PD ({\em Bottleneck})} & 99.90 & 99.90 & 99.90 & 99.20 & 99.40 & 98.60 & 97.10 & 96.90 & 94.30 & 92.70 & 97.79 & 36.00\\
      \hline
      {PI ($L_1$)} &  100.00 & 100.00  & 100.00  & 99.70  & 98.10  & 93.70  & 83.20  & 68.30  & 56.00  & 44.90  &  84.39 & 0.31 \\
      \hline
      {PI ($L_2$)} & 99.90  & 99.50  & 98.60  & 97.40  & 93.10  & 88.50  & 82.90  & 69.70  & 59.40  & 49.90  &  83.89 & 0.26 \\
      \hline
      {PI ($L_\infty$)} & 89.10  & 83.00  & 80.20  & 78.90  &  78.40 & 69.90  & 68.60  & 64.00  & 61.90  & 56.80  &  73.08 & 0.12 \\
      \hline
      {PL ($L_1$)} & 99.20  & 99.70  & 99.00  & 98.50  & 98.50  & 97.30  & 95.90  & 92.30  & 89.10  & 84.50  &  95.40 & 0.74 \\
      \hline
      {PL ($L_2$)} & 99.10  & 99.70  & 98.90  & 98.50  & 98.30  & 96.90  & 95.60  & 92.10  & 89.00  & 84.30  &  95.24 & 0.76 \\
      \hline
      {PL ($L_\infty$)} & 98.90  & 99.60  & 98.80 & 98.40  & 98.30  & 96.50  & 94.80  & 91.70  & 88.70  & 83.80  & 94.95 & 0.09 \\
      \hline
      {PSSK - SVM} & 100.00 & 100.00  & 100.00  & 100.00  & 100.00  & 100.00  & 91.60  & 90.00  & 89.80  &  89.00 &  96.04 & 4.55 \\
      \hline
      {PWGK - SVM} & 100.00 & 100.00 & 100.00 & 100.00 & 100.00 & 99.90 & 99.40 & 95.90 & 87.50 & 73.30 & 95.60 & 0.17 \\
      \hline
      \textbf{PTS \boldmath$(d_\mathbb{G})$} & 100.00 & 100.00 & 100.00 & 100.00 & 100.00 & 99.90 & 99.80 & 98.80 & 96.80 & 93.60 & \textbf{98.89} & \textbf{2.30}\\
      \hline
      \textbf{PTS \boldmath$(d_\Delta)$} & 100.00 & 100.00 & 100.00 & 100.00 & 100.00 & 99.90 & 99.90 & 99.30 & 97.10 & 94.10 & \textbf{99.03} & \textbf{1.60}\\
      \hline
    \end{tabular}} 
    \caption{Comparison of $1$-{\em Wasserstein}, $2$-{\em Wasserstein}, {\em Bottleneck}, $d_\Delta$ and $d_\mathbb{G}$ methods for correctly classifying the topological representations of noisy shapes to their original shape.} \label{table_syn2}
\end{table*} 

%\begin{comment}
\setlength{\intextsep}{0pt}%
\begin{wrapfigure}{r}{0.5\textwidth}\centering
	\includegraphics[width=0.5\columnwidth]{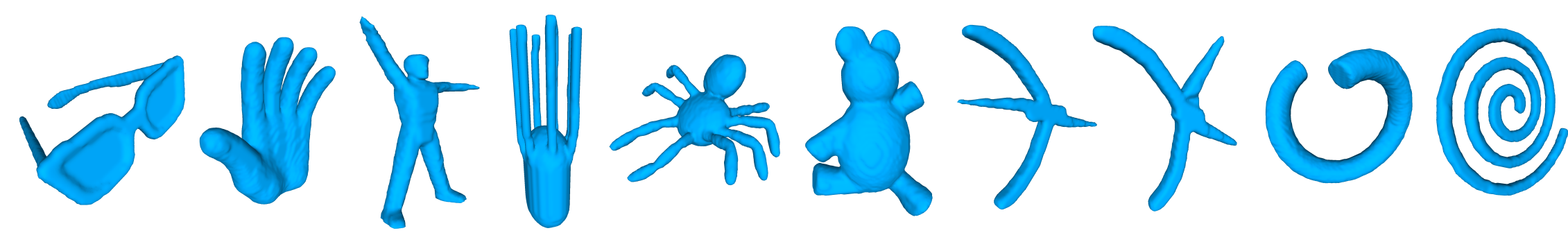}
	\caption{Sample SHREC 2010 shapes used to test robustness of PTS feature to topological noise.}\label{syn_exp_shapes}
\end{wrapfigure}
%\end{comment}

A 17-dimensional scale-invariant heat kernel signature (SIHKS) spectral descriptor function is calculated on each shape \cite{kokkinos2012dense}, and PDs are extracted for each dimension of this function resulting in 17 PDs per shape. The PDs are passed through the proposed framework to get the respective PTS descriptors. The 3D mesh, PD and PTS representation for 4 of the 10 shapes (shown in figure \ref{syn_exp_shapes}) and their respective noisy-variants (Gaussian noise with standard deviation 1.0) is shown in figure \ref{Synthetic_experiment_fig}. In this experiment, we evaluate the robustness of our proposed feature by correctly classifying shapes with different levels of topological noise. Displacement of vertices by adding varying levels of topological noise, interclass similarities and intraclass variations of the shapes make this a challenging task. A simple unbiased one nearest neighbor (1-NN) classifier is used to classify the topological representations of the noisy shapes in each set. The classification results are averaged over the 100 sets and tabulated in table \ref{table_syn2}. We also compare our  method to other TDA-ML methods like PI \cite{adams2017persistence}, PL \cite{bubenik2015statistical}, PSSK \cite{reininghaus2015stable} and PWGK \cite{kusano2016persistence}. For PTS, we set the discretization of the grid $k=50$. For PIs we chose the linear ramp weighting function, set $k$ and $\sigma$ for the Gaussian kernel function, same as our PTS feature. For PLs we use the first landscape function with 500 elements. A linear SVM classifier is used instead of the 1-NN classifier for the PSSK and PWGK methods. From table \ref{table_syn2}, the $2$-\textit{Wasserstein} and {\em Bottleneck} distances over PDs perform poorly even at low levels of topological noise. However, PDs with $1$-\textit{Wasserstein} distance and PTS representations with $d_\mathbb{G}$, $d_\Delta$ metrics show stability and robustness to even high noise levels. {\bf\em Nevertheless, the average time taken to compare two PTS features using either \boldmath{$d_\mathbb{G}$} or \boldmath{$d_\Delta$} is at least two orders of magnitude faster than the \boldmath{$1$}-Wasserstein distance as seen in table \ref{table_syn2}. We also observe that comparison of PIs, PLs and PWGK is an order of magnitude faster than comparing PTS features. However, these methods show significantly lower performance compared to the proposed feature, at correctly classifying noisy shapes as the noise level increases.}

%%%%%%%%%%%%%%%%%%%%%%%%%%%%%%%
\subsection{3D Shape Retrieval} \label{experiments_section_shrec10}

In this experiment, we consider all 10 classes consisting of 200 shapes from the SHREC 2010 dataset, and extract PDs  using 3 different spectral descriptor functions defined on each shape, namely: heat kernel signature (HKS) \cite{sun2009concise}, wave kernel signature (WKS) \cite{aubry2011wave}, and SIHKS \cite{kokkinos2012dense}. HKS and WKS are used to capture the microscopic and macroscopic properties of the 3D mesh surface, while SIHKS descriptor is the scale-invariant version of HKS. 

Using the PTS descriptor we attempt to encode invariances to shape articulations such as rotation, stretching, skewing. For the task of 3D shape retrieval we use a 1-NN classifier to evaluate the performance of the PTS representation against other methods \cite{bubenik2015statistical,reininghaus2015stable,adams2017persistence,li2014persistence,kusano2016persistence}. A linear SVM classifier is used to report the classification accuracy of the PSSK and PWGK methods. Li \textit{et al.} report best results after carefully selecting weights to normalize the distance combinations of their BoF+PD and ISPM+PD methods. As in  \cite{li2014persistence}, we also use the three spectral descriptors and combine our PTS  representations for each descriptor. PIs, PLs and PTS features are also designed the same way as described before. The results reported in table \ref{table_SHREC10} show that the PTS feature (with subspace dimension $p=1$) alone using the $d_{\Delta}$ metric achieves an accuracy of 99.50 \%, outperforming other methods. The average classification result of the PTS feature on varying the subspace dimension $p=1,2,\dots ,25$ is 98.42$\pm$0.4 \% and 98.72$\pm$0.25 \% using $d_{\Delta}$ and $d_{\mathbb{G}}$ metrics respectively, thus displaying its stability with respect to the choice of $p$.

\setlength{\intextsep}{5pt}%
\begin{table*}[htb!]
  \centering
    \scalebox{0.515}{
    \begin{tabular}{|c|c|c|c|c|c|c|c|c|c|c|c|c|c|c|c|c|c|c|}
      \hline
      \textbf{Method} & \makecell{BoF \\ \cite{li2014persistence}} & \makecell{SSBoF \\ \cite{li2014persistence}} & \makecell{ISPM \\ \cite{li2014persistence}} & \makecell{PD \\ ({\em Bottleneck}) \\ \cite{li2014persistence}} & \makecell{PD \\ (1-{\em Wasserstein})} & \makecell{PD \\ (2-{\em Wasserstein})} & \makecell{BoF+PD \\ \cite{li2014persistence}} & \makecell{ISPM+PD \\ \cite{li2014persistence}} & \makecell{PI \\ ($L_1$) \\ \cite{adams2017persistence}} & \makecell{PI \\ ($L_2$) \\ \cite{adams2017persistence}}  & \makecell{PI \\ ($L_\infty$) \\ \cite{adams2017persistence}} & \makecell{PL \\ ($L_1$) \\ \cite{bubenik2015statistical}} & \makecell{PL \\ ($L_2$) \\ \cite{bubenik2015statistical}} & \makecell{PL \\ ($L_\infty$) \\ \cite{bubenik2015statistical}} & \makecell{PSSK \\ (SVM) \\ \cite{reininghaus2015stable}} & \makecell{PWGK \\ (SVM) \\ \cite{kusano2016persistence}} & \makecell{\textbf{PTS} \\ \boldmath$(d_\mathbb{G})$} & \makecell{\textbf{PTS} \\ \boldmath$(d_{\Delta})$} \\
      \hline
      \makecell{\textbf{1-NN} \\ \textbf{Accuracy} \\ \textbf{(\%)}} & 97.00 & 97.50 & 97.50 & 98.50 & 98.50 & 98.50 & 98.50 & 99.00 & 88.50 & 87.50 & 89.50 & 95.00 & 95.00 & 95.00 & 98.50 & 99.00 & \textbf{99.00} & \textbf{99.50}\\
      \hline
		\end{tabular}} 
    \caption{Comparison of the classification performance of the proposed PTS descriptor with other baseline methods \cite{li2014persistence} on the SHREC 2010 dataset.}
\label{table_SHREC10}
\end{table*}

%%%%%%%%%%%%%%%%%%%%%%%%%%%%%
\subsection{View-invariant Activity Analysis} \label{experiments_section_ixmas} 
\setlength{\intextsep}{0pt}%

%\begin{comment}
\begin{wrapfigure}[18]{R}{0.45\textwidth}
\small
\centering
\includegraphics[width=0.45\columnwidth]{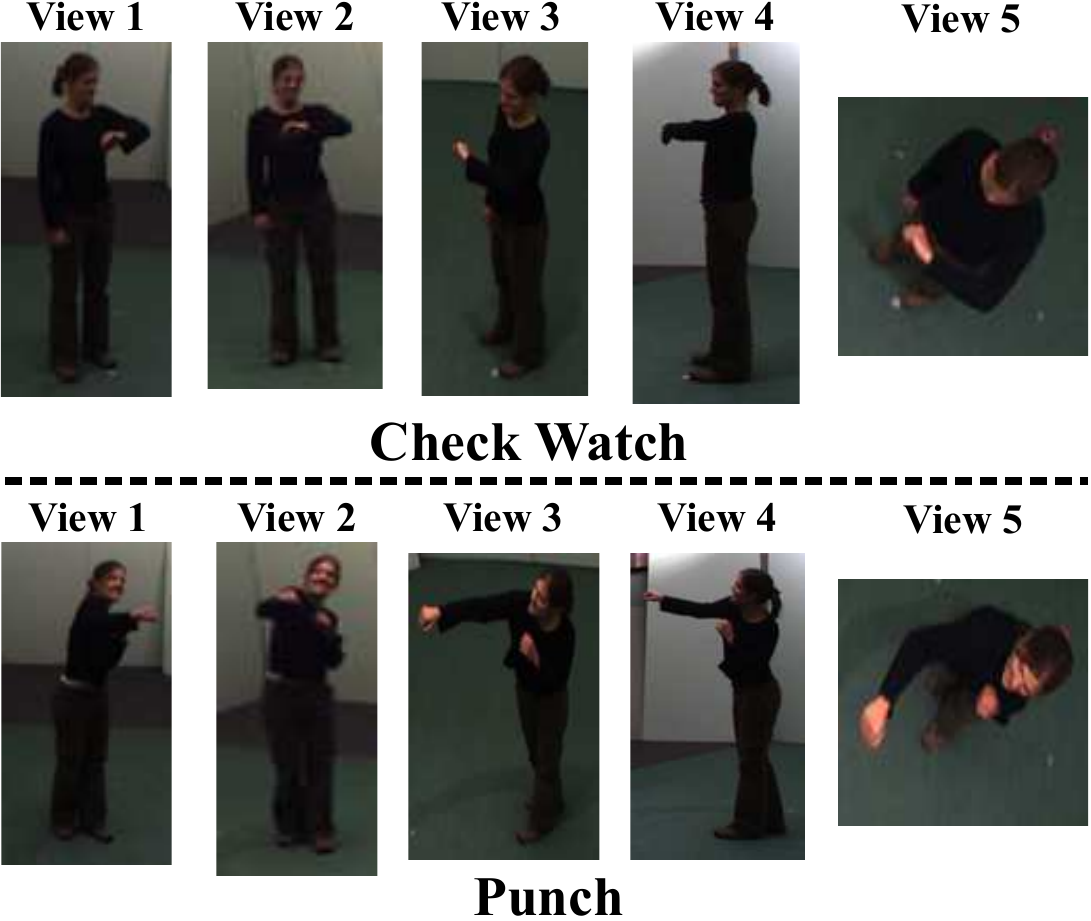}
	\caption{Sample frames for ``check watch'' and ``punch'' action sequences from five views in the IXMAS dataset.}
   	\label{ixmas_frames}
\end{wrapfigure}
%\end{comment}

The IXMAS dataset contains video and silhouette sequences of 11 action classes, performed 3 times by 10 subjects from five different camera views. The 11 classes are as follows - \textit{check watch, cross arms, scratch head, sit down, get up, turn around, walk, wave, punch, kick, pick up}. Sample frames across 5 views for 2 actions are shown in figure \ref{ixmas_frames}. 
We consider only the silhouette information in the dataset for our PTS representations. For each frame in an action sequence we extract multi-scale shape distributions which are referred to as A3M, D1M, D2M and D3M, over the 2D silhouettes \cite{som2017multiscale}. The multi-scale shape distribution feature captures the local to global changes in different geometric properties of a shape. For additional details about this feature, please see: \cite{som2017multiscale,som2016attractor,osada2002shape}. 

For $n$  frames in an action sequence and $b$ bins in each shape distribution at a certain scale, an $n\times b$ matrix representing the action is obtained. Treating the $n$ frames as nodes, scalar field topological PDs are calculated across each column, resulting in $b$ PDs. PDs capture the structural changes along each bin in the distributions. We select 5 different scales for the multi-scale shape features, giving us $5b$ PDs per action which are passed through the proposed pipeline resulting in $5b$ PTS features. PTS features try to encode the possible changes with respect to view-point variation, body-type and execution style. To represent the entire action as a point on the Grassmannian, we select the first two largest singular vectors from each of the $5b$ PTS descriptors, apply SVD and choose 20 largest components.

\setlength{\intextsep}{0pt}%
\begin{wraptable}{r}{0.55\textwidth}
\centering
\small
\scalebox{0.635}{
\begin{tabular}{|c|c|c|c|c|}
\hline
\multirow{2}{*}{ \textbf{Method}} & \multicolumn{2}{|c|}{ \makecell{\textbf{Same Camera}\\ \textbf{Accuracy (\%)}}} & \multicolumn{2}{|c|}{ \makecell{\textbf{Any-To-Any}\\ \textbf{Accuracy (\%)}}} \\ 
\cline{2-5}
& \textbf{Best} & \textbf{Mean}$\bm{\pm}$\textbf{SD} & \textbf{Best} & \textbf{Mean}$\bm{\pm}$\textbf{SD} \\
\hline

%% Other baselines 
%\hline

%% Baseline
SSM-HOG \cite{junejo2011view} & 67.30 & - & 52.60 & -\\
PTS-HOG & 51.31 & - & 41.24 & - \\
SSM-HOG + PTS-HOG & 69.01 & - & 55.13 & - \\

%% Baseline + Proposed
SSM-HOG + PTS-A3M & 73.15 & 72.06$\bm{\pm}$1.14 & 58.36 & 56.96$\bm{\pm}$1.05 \\
SSM-HOG + PTS-D1M & 74.25 & 73.26$\bm{\pm}$1.53 & 59.26 & 57.67$\bm{\pm}$1.19 \\
SSM-HOG + PTS-D2M & 74.92 & 74.22$\bm{\pm}$1.36 & 59.77 & 58.19$\bm{\pm}$1.03 \\
\textbf{SSM-HOG + PTS-D3M} & \textbf{76.18} & 73.72$\bm{\pm}$1.13 & \textbf{60.33} & 58.72$\bm{\pm}$1.11 \\
\hline

%% Baseline
SSM-OF \cite{junejo2011view} & 66.60 & - & 53.80 & - \\
%% Baseline + Proposed
SSM-OF + PTS-A3M & 72.02 & 70.25$\bm{\pm}$1.06 & 58.85 & 57.48$\bm{\pm}$0.93 \\
SSM-OF + PTS-D1M & 73.67 & 71.62$\bm{\pm}$1.17 & 59.56 & 57.81$\bm{\pm}$1.05 \\
SSM-OF + PTS-D2M & 73.45 & 72.53$\bm{\pm}$1.12 &  60.60 & 59.05$\bm{\pm}$1.11 \\
\textbf{SSM-OF + PTS-D3M} & \textbf{74.41} & 72.21$\bm{\pm}$1.03 & \textbf{61.51}& 59.33$\bm{\pm}$1.13 \\
\hline

%% Baseline
SSM-HOG-OF \cite{junejo2011view} & 76.28 & - & 61.25 & - \\
%% Baseline + Proposed
SSM-HOG-OF + PTS-A3M & 79.30 & 78.05$\bm{\pm}$0.71 & 64.93 & 63.58$\bm{\pm}$0.65 \\
SSM-HOG-OF + PTS-D1M & 79.61 & 79.03$\bm{\pm}$0.96 & 65.39 & 64.27$\bm{\pm}$0.65 \\
SSM-HOG-OF + PTS-D2M & 79.86 & 79.35$\bm{\pm}$0.76 & 65.70 & 64.62$\bm{\pm}$0.83 \\
\textbf{SSM-HOG-OF + PTS-D3M} & \textbf{81.12} & 79.49$\bm{\pm}$0.99 & \textbf{66.16} & 64.99$\bm{\pm}$0.79 \\
\hline
\end{tabular}}
\caption{Comparison of the recognition results on the IXMAS dataset. Results are presented for two combinations of train camera \textit{X} and test camera \textit{Y}. ``Same Camera" denotes \textit{X=Y}; ``Any-To-Any'' implies any combination of \textit{X},\textit{Y}.}
\label{table2_IXMAS}
\end{wraptable}

To perform multi-view action recognition, we train non-linear SVMs using the Grassmannian RBF kernel, $k_{rp}(\mathcal{X}_i,\mathcal{Y}_i) = \textrm{exp} \Big( - \beta \|{\mathcal{X}_i}^{\textrm{T}}\mathcal{Y}_i{\|^2_F} \Big), \ $ $\beta>0$ \cite{harandi2014expanding}. Here, $\mathcal{X}_i$, $\mathcal{Y}_i$ are points on the Grassmannian and $\|.\|_F$ is the Frobenius norm. We set $\beta=1$ in our implementations. Junejo \textit{et al.} train non-linear SVMs using the $\chi^2$ kernel over the SSM-based descriptors and follow a one-against-all approach for multi-class classification \cite{junejo2011view}. We follow the same approach and use a joint weighted kernel between their SSM kernel and our kernel, \textit{i.e.} $\chi^2 + \lambda \cdot k_{rp}$, where $\lambda = 0.1,0.2,\dots 1.0$. The SSM-based descriptors are computed using the histogram of gradients (HOG), optical flow (OF) and fusion of HOG, OF features. The classification results are tabulated in table \ref{table2_IXMAS}. Apart from reporting results of PTS representations obtained using the multi-scale shape distributions, we also show recognition results of PTS feature computed over the HOG descriptor (PTS-HOG). We see significant improvement in the results by fusing different PTS features with the SSM-based descriptor. We also tabulate the mean and standard deviation values for all classification results obtained after varying $\lambda$ from 0.1 to 1.0 and subspace dimension $p$ from 1 to 10. These results demonstrate the flexibility and stability associated with the proposed PTS topological descriptor.

%%%%%%%%%%%%%%%%%%%%%%%%%%%%%%%
\subsection{Dynamical Analysis on Motion Capture Data} \label{experiments_section_mocap}

\setlength{\intextsep}{0pt}%
\begin{wraptable}[16]{r}{0.445\textwidth}
\centering
\small
\scalebox{0.8}{
\begin{tabular}{|c|c|c|}
\hline
\textbf{Method} & \makecell{\textbf{Accuracy} \\ \textbf{(\%)}} & \makecell{\textbf{Average} \\ \textbf{Time Taken} \\ \textbf{(}$\bm{10^{-4}}$ \textbf{sec)}}\\ 
\hline

%% Baseline
%Chaos \cite{ali2007chaotic} & 52.44 & -\\
\makecell{PD ($1$-Wasserstein)\\NN \cite{zomorodian2010fast}} & 93.68 & 22.00\\
%T-VR Complex - NN \cite{venkataraman2016persistent} & 96.48 & 1200 \\
\hline
\makecell{Hilbert Sphere\\ NN \cite{anirudh2016riemannian}} & 89.87 & 590.00\\
\makecell{Hilbert Sphere \\ PGA+SVM \cite{anirudh2016riemannian}} & 91.68 & -\\
\hline

%% Proposed
\textbf{PTS \boldmath$(d_\Delta)$ - NN} & 85.96 & 0.19\\
\textbf{PTS - SVM} & 91.92 & -\\
\hline

\end{tabular}}
\caption{Comparison of classification performance and the average time taken to compare two feature representations on the motion capture dataset.}
\label{table_MoCap}
\end{wraptable}

This dataset consists of human body joint motion capture sequences in 3D, where each sequence contains 57 trajectories (19 joint trajectories along 3 axes). There are 5 action classes - \textit{dance, jump, run, sit and walk}, with each class containing 31, 14, 30, 35 and 48 sequences respectively. $H_1$ homology group PDs are computed over the reconstructed attractor for each trajectory, resulting in 57 PDs per action \cite{anirudh2016riemannian} and the corresponding PTS feature is also extracted. We report the average classification performance over 100 random splits, with each split having 25 random test samples (5 samples from each class) and remaining 133 training samples. For SVM classification, we train non-linear SVMs using the projection kernel, $k_p(\mathcal{X}_i,\mathcal{Y}_i) = \|{\mathcal{X}_i}^{\textrm{T}}\mathcal{Y}_i{\|^2_F}$ \cite{hamm2008grassmann}. 

The results are tabulated in table \ref{table_MoCap}. PTS features have a classification accuracy of 85.96 \% and 91.92 \% using the 1-NN and SVM classifier respectively. While these results are slightly lower than the $1$-\textit{Wasserstein} metric, the proposed descriptor with the $d_{\Delta}$ metric is more than 2 orders of magnitude faster. Topological properties of dynamic attractors for analysis of time-series data has been studied and applied to tasks such as wheeze detection \cite{DBLP:journals/spl/EmraniSMK15}, pulse pressure wave analysis \cite{DBLP:journals/spl/EmraniGK14} and such applications are surveyed in \cite{DBLP:journals/spm/KrimGC16}. We ask our readers to refer to these papers for further exploration.

%%%%%%%%%%%%%%%%%%%%%%%%%%%%%%%
\subsection{Time-complexity of Comparing Topological Representations} 

\begin{table*}[b!]
\centering
\small
\scalebox{0.585}{    
\begin{tabular}{|c|c|c|c|c|c|c|c|}
\hline
\multirow{2}{*}{ \textbf{Dataset}} & \multirow{2}{*}{ \makecell{\textbf{Average Number} \\ \textbf{of Points in PD}}} &  \multicolumn{5}{|c|}{ \textbf{Average Time Taken \textbf{(}$\bm{10^{-4}}$ \textbf{sec)}}} & \multirow{2}{*}{ \makecell{\textbf{Subspace Dimension ($p$)} \\ \textbf{of PTS Feature}}} \\
\cline{3-7}
&& \boldmath$1$\textbf{-Wasserstein} & \boldmath$2$\textbf{-Wasserstein} & \textbf{Bottleneck} & \boldmath$d_\mathbb{G}$ & \boldmath$d_\Delta$ &\\
\hline
\textbf{SHREC 2010 \cite{lian2010shrec}} & 71 & \makecell{256.00 \\ (Kerber \textit{et al.} \cite{kerber2017geometry}: \\ 219.00)} & \makecell{450.00 \\ (Kerber \textit{et al.} \cite{kerber2017geometry}:\\ 237.00)} & \makecell{36.00 \\ (Kerber \textit{et al.} \cite{kerber2017geometry}: \\ 295.00)} & \textbf{2.30} & \textbf{1.60} & 10\\
\hline
\textbf{IXMAS \cite{weinland2007action}} & 23 & 16.00 & 16.00 & 3.43 & \textbf{2.21} & \textbf{0.68} & 20\\
\hline
\textbf{Motion Capture \cite{ali2007chaotic}} & 27 & 22.00 & 22.00 & 2.72 & \textbf{0.24} & \textbf{0.19} & 1\\
\hline
\end{tabular}}
\caption{Comparison of the average time taken to measure distance between two PDs using the $1$-Wasserstein, $2$-Wasserstein and Bottleneck metrics, and between two PTS features using $d_\mathbb{G}$ and $d_\Delta$ metrics. The time reported is averaged over 3000 distance calculations between the respective topological representations for all three datasets used in section \ref{experiments_section}.} \label{time_table}
\end{table*}

All experiments are carried out on a standard Intel i7 CPU  using Matlab 2016b with a working memory of 32 GB. We used the Hungarian algorithm to compute the \textit{Bottleneck} and $p$-\textit{Wasserstein} distances between PDs. Kerber \textit{et al.} take advantage of the geometric structure of the input graph and propose geometric variants of the above metrics, thereby showing significant improvements in runtime performance when comparing PDs having several thousand points \cite{kerber2017geometry}. However, extracting PDs for most real datasets of interest in this paper does not result in more than a few hundred points. For example, on average we observe 71, 23, 27 points in each PD for the SHREC 2010, IXMAS and motion capture datasets respectively. The Hungarian algorithm incurs similar computations in this setting as shown in table \ref{time_table}. The $d_{\mathbb{G}}$ and $d_{\Delta}$ metrics used to compare different PTS representations (grid size $k$ = 50) are fast and computationally less complex compared to the {\em Bottleneck} and $p$-{\em Wasserstein} distance measures. The average time taken to compare two topological signatures (PD or PTS) for each of the datasets is tabulated in table \ref{time_table}. The table also shows the average number of points seen per PD and the subspace dimension $p$ used for the PTS representation.

\setlength{\intextsep}{0pt}%
\begin{wraptable}{r}{0.625\textwidth}
\centering
\small
\scalebox{0.725}{    
\begin{tabular}{|c|c|c|c|c|c|c|c|c|c|c|c|}
\hline
 &  \multicolumn{11}{|c|}{ \textbf{Average Time Taken \textbf{(}$\bm{10^{-4}}$ \textbf{sec)}}} \\
\hline
\cline{2-12}
\textbf{Grid size (\textit{k})} & 5 & 10 & 20 & 40 & 60 & 80 & 100 & 200 & 300 & 400 & 500 \\
\hline
\textbf{PTS (\boldmath$d_\mathbb{G}$)} & 0.72 & 0.73 & 0.89 & 1.31 & 1.48 & 2.28 & 5.53 & 8.35 & 18.40 & 32.88 & 47.07\\
\hline
\textbf{PTS (\boldmath$d_\Delta$)} & 0.20 & 0.33 & 0.84 & 0.72 & 1.00 & 1.85 & 4.32 & 7.70 & 17.69 & 31.56 & 46.68\\
\hline
\end{tabular}}
\caption{Comparison of the average time taken to measure distance between two PTS features using $d_\mathbb{G}$ and $d_\Delta$ metrics w.r.t. variation in grid size $k$. The time reported is averaged over 3000 distance calculations between the topological representations for the SHREC 2010 dataset.} \label{time_table_varying_grid}
\end{wraptable}

Table \ref{time_table_varying_grid} shows the variation of the average time taken to compare PTS features on varying the grid size ($k$) of the 2D PDF. Here too the average time is reported after averaging over 3000 distance calculations between PTS features computed from PDs of the SHREC 2010 dataset. We observe that the time taken to compare two PTS features with a grid size $k = 500$ is two orders of magnitude greater than the time obtained for PTS features using $k = 5$. However, these times are still smaller than or on par with the times reported using $p$-\textit{Wasserstein} and \textit{Bottleneck} distances between PDs as seen in table \ref{time_table}. For all our experiments we set $k=50$ for our PTS representations and as shown in table \ref{time_table}, the times reported for $d_\Delta$ and $d_\mathbb{G}$ are at least an order of magnitude faster than \textit{Bottleneck} distance and two orders of magnitude faster than the $p$-\textit{Wasserstein} metrics.

%%%%%%%%%%%%%%%%%%%%%%%%%%%%%%%
\section{Conclusion and Discussion} \label{conclusion_section}
%%%%%%%%%%%%%%%%%%%%%%%%%%%%%%%

We believe that a perturbed realization of a PD computed over a high-dimensional shape/graph is robust to topological noise affecting the original shape. Based on the type of data and application, topological noise can imply different types of variations, such as: articulation in 3D shape point cloud data; diversity in body structure, execution style and view-point pertaining to human actions in video analysis, \textit{etc}. In this paper, we propose a novel topological representations called PTS that is obtained using a perturbation approach, taking first steps towards robust invariant learning with topological features. 
We obtained perturbed persistence surfaces and summarized them as a point on the Grassmann manifold, in order to utilize the different distance metrics and Mercer kernels defined for the Grassmannian. The $d_\mathbb{G}$ and $d_\Delta$ metrics used to compare different Grassmann representations are computationally cheap as they do not depend on the number of points present in the PD, in contrast to {\em Bottleneck} and $p$-{\em Wasserstein} metrics, which do. The PTS feature offers flexibility in choosing the weighting function, kernel function and perturbation level. This makes it easily adaptable to different types of real-world data. It can also be easily integrated with various ML tools, which is not easily achievable with PDs. 
Future directions include fusion with contemporary deep-learning architectures to exploit the complementarity of both paradigms. We expect that topological methods will push the state-of-the-art in invariant representations, where the requisite invariance is incorporated using a topological property of an appropriately redefined metric space. Additionally, the proposed methods may help open new feature-pooling options in deep-nets.

\clearpage


\bibliographystyle{splncs04}
\bibliography{egbib}

\end{document}